\documentclass[11pt]{article}

\usepackage[preprint]{acl}

\usepackage{times}
\usepackage{latexsym}

\usepackage[T1]{fontenc}
\usepackage[utf8]{inputenc}
\usepackage{microtype}
\usepackage{inconsolata}
\usepackage{graphicx}
\usepackage{booktabs}
\usepackage{multirow}
\usepackage{array}
\usepackage{amsmath}
\usepackage{amssymb}
\usepackage{enumitem}
\usepackage{makecell}
\usepackage[table]{xcolor}
\usepackage{tcolorbox}
\tcbuselibrary{skins, breakable}
\usepackage{fontawesome5}
\usepackage{longtable}
\usepackage{hyperref}

\newcommand{\eg}{{\it e.g.}}

\newcommand{\ie}{{\it i.e.}}

\title{PACE-Bench: Benchmarking \underline{\textit{P}}hysics \underline{\textit{A}}daptation via \underline{\textit{C}}ode \underline{\textit{E}}volution \\ in Dynamic Environments}

\author{Yuhao Zhan\textsuperscript{1,2}\thanks{indicates equal contribution.}, Bingxiang He\textsuperscript{1}$^*$, Zecong Tang\textsuperscript{2}, Chaojun Xiao\textsuperscript{1}\thanks{indicates corresponding author.} \\
  \textsuperscript{1} Tsinghua University\\
  \textsuperscript{2}Zhejiang University \\
  \texttt{yuhao.zhan@zju.edu.cn}
}

\begin{document}

\newcommand{\greenc}{\textcolor{green!40!black}{\checkmark}}
\newcommand{\redx}{\textcolor{red!60!black}{\textbf{$\times$}}}
\newcommand{\yellowt}{\textcolor{yellow!70!black}{$\triangle$}}
\newcommand{\greencheck}{\textcolor{green!40!black}{\checkmark}}

\newcommand{\TODO}[1]{{\color{red}\textbf{[TODO: #1]}}}

\maketitle

\begin{abstract}
Self-evolving agents improve future behavior from interaction experience, yet existing evaluations typically optimize under fixed execution conditions and do not test recovery after those conditions change.
To address this gap, we introduce \textsc{PACE-Bench} (\textbf{P}hysics \textbf{A}daptation via \textbf{C}ode \textbf{E}volution), a simulator-grounded benchmark of 144 source-to-target adaptation pairs across six physics domains. Each pair links a source environment to a mutated target environment with the same goal and interface. A code-driven design that succeeds in the source fails in the target, where agents must iteratively adapt it into a working target design using diagnostic sandbox feedback within a limited attempt budget. We compare ten self-evolving methods from four paradigms. The benchmark remains far from saturated: Reflexion + Qwen3-14B succeeds on only 35.9\% of full-benchmark pairs, while GPT-5.5 solves 66.7\% of the Statics subset under the full budget. Together, these results show that simulator-grounded reflection is more reliable than unverified self-revision, while memory anchors agents to early designs and broad tree search explores without converging. Even revealing exact physical changes does not raise the performance ceiling, pointing to mechanism redesign rather than parameter inference as the central bottleneck. Data and code are available at \url{https://github.com/thunlp/PACE-Bench}.
  \end{abstract}

\section{Introduction}

\begin{figure}[t]
  \centering
  \includegraphics[width=\columnwidth]{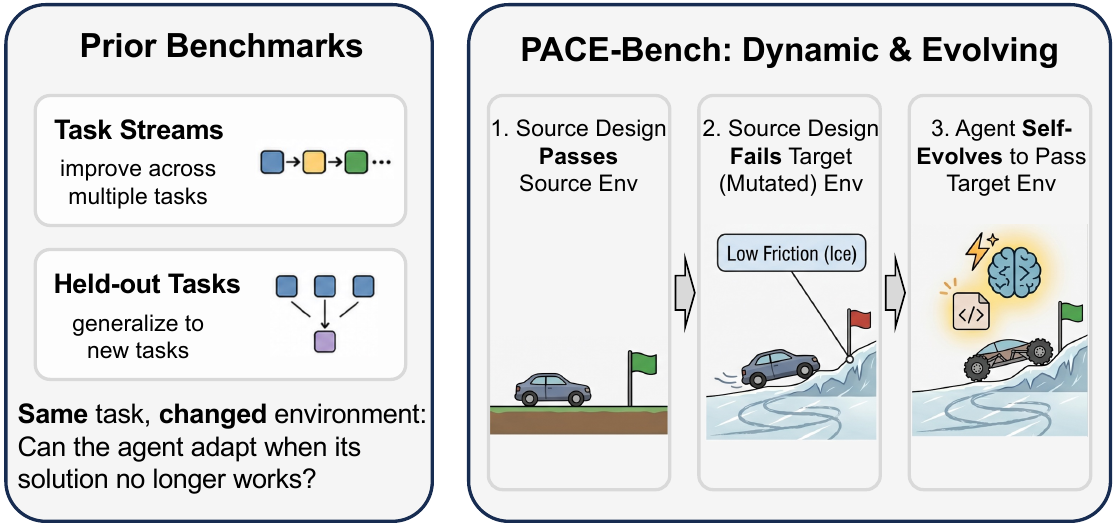}
  \caption{Existing self-evolving benchmarks test transfer or generalization, whereas \textsc{PACE-Bench} tests adaptation of an executable design after physical environment shift.}
  \label{fig:comparison}
  \vspace{-0.5cm}
\end{figure}

\begin{table*}[t]
  \centering
  \resizebox{0.98\textwidth}{!}{%
  \begin{tabular}{lcccc>{\columncolor{blue!8}}c}
    \toprule
    \textbf{Benchmark}
      & \makecell{\textbf{Feedback-Driven}\\ \textbf{Improvement}}
      & \makecell{\textbf{Physical} \\ \textbf{Simulation}}
      & \makecell{\textbf{Executable} \\ \textbf{Design}}
      & \makecell{\textbf{Self-Evolution} \\ \textbf{Evaluation}}
      & \makecell{\textcolor{blue!60!black}{\textbf{Adaptation After}}\\ \textcolor{blue!60!black}{\textbf{Env. Change}}} \\
    \midrule
    \multicolumn{6}{l}{\textit{Self-evolving agent benchmarks}} \\
    StreamBench \cite{wu2024streambenchbenchmarkingcontinuousimprovement} & \greenc & \redx & \redx & \greenc & \redx \\
    LifelongAgentBench \cite{zheng2025lifelongagentbenchevaluatingllmagents} & \greenc & \redx & \redx & \greenc & \redx \\
    SE-Bench \cite{yuan2026sebenchbenchmarkingselfevolutionknowledge} & \redx & \redx & \redx & \greenc & \redx \\
    SEA-Eval \cite{jiang2026seaevalbenchmarkevaluatingselfevolving} & \greenc & \redx & \redx & \greenc & \redx \\
    SkillLearnBench \cite{zhong2026skilllearnbenchbenchmarkingcontinuallearning} & \greenc & \redx & \redx & \greenc & \redx \\
    SEAGym \cite{zheng2026seagymevaluationenvironmentselfevolving} & \greenc & \redx & \redx & \greenc & \redx \\
    AutoEnv \cite{zhang2025autoenvautomatedenvironmentsmeasuring} & \greenc & \redx & \redx & \greenc & \redx \\
    EvoAgentBench \cite{gao2026evoagentbenchbenchmarkingagentselfevolution} & \greenc & \redx & \redx & \greenc & \redx \\
    \midrule
    \multicolumn{6}{l}{\textit{Physical and engineering benchmarks}} \\
    ENGDESIGN \cite{guo2026toward} & \redx & \greenc & \greenc & \redx & \redx \\
    Frontier-Eng \cite{chi2026frontier} & \greenc & \greenc & \greenc & \greenc & \redx \\
    CausalWorld \cite{ahmed2020causalworld} & \greenc & \greenc & \redx & \redx & \redx \\
    NewtonBench \cite{zheng2025newtonbench} & \greenc & \greenc & \redx & \redx & \redx \\
    \midrule
    \textbf{PACE-Bench} & \greenc & \greenc & \greenc & \greenc & \greenc \\
    \bottomrule
  \end{tabular}%
  }
  \caption{Comparison of representative benchmarks for self-evolution, engineering design, and physical environments. The highlighted final column is the core capability evaluated by \textsc{PACE-Bench}; the other columns describe its evaluation setting.}
  \label{tab:comparison}
  \vspace{-0.2cm}
\end{table*}

Self-evolving agents improve future behavior by updating their parameters, context, memory, or tools from interaction experience \cite{gao2025survey}. By turning interaction feedback into autonomous improvement, this paradigm reduces repeated human redesign and supports scalable, lifelong agent systems \cite{fang2025comprehensivesurveyselfevolvingai}.

Despite this promise, standard evaluations optimize an answer or policy under fixed execution rules. Recent benchmarks measure experience accumulation over task streams \cite{wu2024streambenchbenchmarkingcontinuousimprovement,zheng2025lifelongagentbenchevaluatingllmagents}, including long-term performance, efficiency, and stability \cite{jiang2026seaevalbenchmarkevaluatingselfevolving}, or transfer learned knowledge and skills to related tasks \cite{yuan2026sebenchbenchmarkingselfevolutionknowledge,zhong2026skilllearnbenchbenchmarkingcontinuallearning,gao2026evoagentbenchbenchmarkingagentselfevolution}. Others test whether prompt, code, or harness updates generalize to held-out tasks and environments \cite{zheng2026seagymevaluationenvironmentselfevolving,zhang2025autoenvautomatedenvironmentsmeasuring}. However, none evaluates whether an agent can adapt a previously successful design after a change in the execution environment causes it to fail. Such shifts are fundamental in deployment, where surroundings and operating conditions evolve frequently. Agents must identify obsolete assumptions, preserve useful components, and decide whether to revise or replace the design.

Physical simulation offers a controlled testbed for this adaptation, while code provides an executable interface for revising and verifying designs. Changes in friction, material strength, or dynamics can invalidate a working design without changing its goal. Figure~\ref{fig:comparison} illustrates this with a vehicle that works on high-friction terrain but fails on ice. Existing physical benchmarks, however, either study simulator-based design or executable feedback under fixed settings \cite{guo2026toward,chi2026frontier}, or vary physical environments without self-evolving redesign \cite{ahmed2020causalworld,zheng2025newtonbench}. Adaptation of an existing design after an environment change therefore remains untested (Table~\ref{tab:comparison}).

To address this gap, we introduce \textsc{PACE-Bench}, a benchmark of self-evolving adaptation after controlled environment changes using executable physical design. Following prior work on code-based agents \cite{wang2024executablecodeactionselicit} and executable physical simulation \cite{xie2026physcodebenchbenchmarkingphysicsawaresymbolic}, each task starts with a code-driven design that succeeds in a source environment. A \textit{mutation} then systematically changes physical parameters such as friction, material strength, or dynamics to create a target environment where the source design fails but a verified reference design succeeds. With the goal and interface fixed, the agent has 20 attempts to iteratively revise its design using diagnostic sandbox feedback. Across 36 base tasks and six physics domains, four targets per task yield 144 source-to-target pairs ($\tau_0 \to \tau_k$).

We compare ten self-evolving methods from four paradigms across Qwen3 model sizes, then evaluate larger open and frontier LLMs on selected subsets. The benchmark remains far from saturated: Reflexion with Qwen3-14B succeeds on only 35.9\% of full-benchmark pairs, while GPT-5.5 solves 66.7\% of the Statics subset under the full budget. Simulator-grounded reflection outperforms unverified self-revision, memory can anchor agents to early designs, and broad tree search often struggles to convert exploration into convergence. Code-similarity and nine-category error analyses trace these behaviors to Design Fixation versus Stagnation or undirected Exploration. Finally, revealing exact physical changes still does not raise the performance ceiling, indicating that mechanism redesign (``\textit{know how}'') is the genuine bottleneck compared to parameter inference (``\textit{know what}''). Together, these findings position \textsc{PACE-Bench} as a reproducible testbed for diagnosing and improving self-evolving agents under changing environments.

\noindent\textbf{Contributions.} (i) We introduce \textsc{PACE-Bench}, a systematic benchmark for self-evolving adaptation after environment shifts using code-driven physical design, with a unified codebase and reproducible evaluation suite. (ii) We evaluate ten methods across four paradigms, showing that the benchmark remains unsaturated, simulator-grounded reflection is reliable, tree search is efficient, and memory can constrain stronger models. (iii) We analyze failure trajectories and parameter disclosure, revealing the tension between exploration and exploitation and the difficulty of mechanism redesign even when exact physical changes are known.

\section{Related Work}

\noindent\textbf{Self-Evolving Benchmarks and Physical Design.}
Existing self-evolving benchmarks extend evaluation beyond isolated episodes. Some measure whether feedback improves performance over task streams \cite{wu2024streambenchbenchmarkingcontinuousimprovement,zheng2025lifelongagentbenchevaluatingllmagents}, including long-term performance, efficiency, and stability \cite{jiang2026seaevalbenchmarkevaluatingselfevolving}, or whether learned knowledge and skills transfer to related tasks \cite{yuan2026sebenchbenchmarkingselfevolutionknowledge,zhong2026skilllearnbenchbenchmarkingcontinuallearning,gao2026evoagentbenchbenchmarkingagentselfevolution}. Other benchmarks modify prompts, code, or agent harnesses to test generalization to held-out tasks and heterogeneous environments \cite{zheng2026seagymevaluationenvironmentselfevolving,zhang2025autoenvautomatedenvironmentsmeasuring}. None tests whether an agent can adapt its design when the environment changes while the goal and interface stay fixed.
Physical adaptation offers a controlled testbed for this case. Existing work studies simulator-based design \cite{guo2026toward}, executable feedback \cite{chi2026frontier}, causal transfer \cite{ahmed2020causalworld}, or physical-law discovery \cite{zheng2025newtonbench}. Yet none combines a failure after an environment change, code-driven mechanism redesign, and systematic self-evolution evaluation with a verified solvable target. \textsc{PACE-Bench} bridges these settings by comparing self-evolving methods on executable designs after controlled physics shifts (Table~\ref{tab:comparison}).

\noindent\textbf{Self-Evolving Agents.}
A self-evolving agent modifies its parameters, contextual state, or tools based on its own interaction trajectories to improve future performance \cite{gao2025survey}. Recent work spans four paradigms: \textbf{context-based} revision through iterative refinement and self-critique \cite{shinn2023reflexion,madaan2023self}; \textbf{memory-augmented} approaches that maintain structured experience across attempts \cite{zhao2024expel,ouyang2025reasoningbank,zhang2025agentic}; \textbf{inference-time search} over solution candidates via evolutionary algorithms \cite{novikov2025alphaevolve,assumpccao2025codeevolve}; and \textbf{parameter-based} training through test-time RL \cite{wang2025ragen,wang2025thetaevolve,yuksekgonul2026learning}, SFT \cite{zweiger2026self}, genetic search \cite{zhang2025nature}, or self-generated data \cite{zhao2026absolute}. These methods are driven by textual feedback, scalar rewards, or population-based selection. While prior benchmarks rarely evaluate self-evolving methods systematically, \textsc{PACE-Bench} compares them across four paradigms under physical adaptation tasks.

\begin{figure*}[t]
  \centering
  \includegraphics[width=0.85\textwidth]{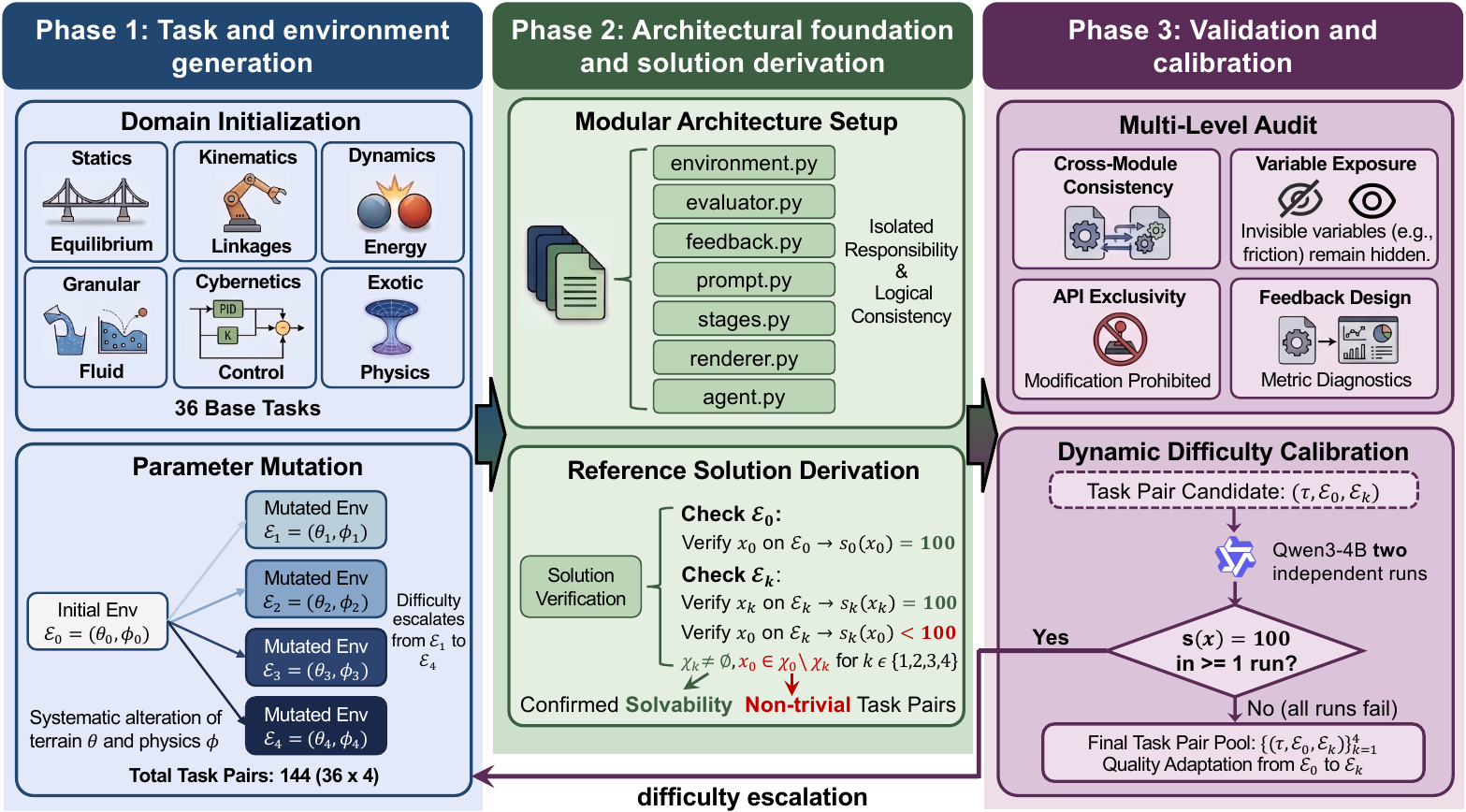}
  \caption{Overview of the three-phase dataset construction pipeline.}
  \label{fig:construction}
  \vspace{-0.3cm}
\end{figure*}

\section{The \textsc{PACE-Bench} Benchmark}

\subsection{Task Formulation}

A \textit{base task} in \textsc{PACE-Bench} is defined by a fixed context template $\mathcal{C}$ and five environments $\{\mathcal{E}_0,\ldots,\mathcal{E}_4\}$. The template specifies the natural-language description, constraints, success criteria, and permitted primitive APIs (\eg, \texttt{sandbox.add\_beam(x, y, width, height)}). Its environment-conditioned is:
\begin{equation}
\tau_i = \mathcal{C}(\mathcal{E}_i),
\end{equation}
where $\mathcal{E}_0$ is the source environment and $\mathcal{E}_1, \dots, \mathcal{E}_4$ (Stages 1--4) are mutated environments with systematically altered physical parameters. Thus, the five environment-conditioned instances share the task goal and interface but differ in physical conditions. We suppress the base-task index below.

Each environment $\mathcal{E}_i = (\theta_i, \phi_i)$ is parameterized by terrain $\theta_i$ and physics $\phi_i$. A \textit{solution} $x$ is a Python program defining \texttt{build\_agent()} (structure assembly) and \texttt{agent\_action()} (control logic). The evaluator executes $x$ in the Box2D\footnote{Box2D is widely used in platforms such as OpenAI Gym \cite{brockman2016openaigym} and CARL \cite{benjamins2021carlbenchmarkcontextualadaptive}.} sandbox and returns structured feedback:
\begin{equation}
E_i(x) = \bigl(v(x),\; s(x),\; d(x)\bigr),
\end{equation}
where $v(x) = K/N$ is the fraction of $N$ hard constraints satisfied (\eg, mass budget, no collapse), and $d(x)$ is a diagnostic report (\eg, peak joint force, failure timestamp). The task score $s(x)$ is:
\begin{equation}
s(x) = \begin{cases}
(v(x) - 1) \times 100, & \text{if } v(x) < 1,\\[4pt]
\text{task progress},   & \text{if } v(x) = 1.
\end{cases}
\end{equation}
When $v(x) < 1$, the score penalizes violated constraints. When $v(x) = 1$, it measures progress toward the task goal. Moreover, $s(x) = 100$ if and only if $x \in \mathcal{X}$, where $\mathcal{X}$ denotes the set of reference solutions that pass the task.

\subsection{Self-Evolving Evaluation Protocol}
\label{sec:eval-protocol}

The core evaluation unit is a \textit{source-to-target environment pair} $(\mathcal{E}_0 \to \mathcal{E}_k)$ within one base task, evaluated through the corresponding task instances $\tau_0 = \mathcal{C}(\mathcal{E}_0)$ and $\tau_k = \mathcal{C}(\mathcal{E}_k)$, where $k \in \{1,2,3,4\}$. Each pair satisfies:
\begin{equation}
\mathcal{X}_k \neq \emptyset \quad\text{and}\quad x_0 \in \mathcal{X}_0 \setminus \mathcal{X}_k,
\end{equation}
where $\mathcal{X}_k \neq \emptyset$ ensures that $\tau_k$ is solvable, and $x_0 \in \mathcal{X}_0 \setminus \mathcal{X}_k$ rules out trivial reuse of the source solution. The agent must adapt $x_0$ to $\tau_k$ without being told which physics changed, using feedback to infer the change and redesign accordingly.

The agent operates under an interaction budget $B = 20$. Each \textit{attempt} submits a candidate $x^t$ and returns feedback $(v^t, s^t, d^t) = E_k(x^t)$. At step $t$, the agent $\mathcal{A}$, an LLM augmented with a self-evolving method, conditions on the task context and prior history:
\begin{equation}
x^t = \mathcal{A}\bigl(\mathcal{C}(\mathcal{E}_k) \oplus \mathcal{U},\; H^{t-1}\bigr),
\end{equation}
\begin{equation}
H^{t-1} = \{(x^i, v^i, s^i, d^i)\}_{i=0}^{t-1},
\end{equation}
where $\mathcal{U}$ is a Uniform Suffix (see~\S\ref{sec:module-audit}) listing variables that \textit{might} differ across stages without identifying the changed variables. The final task score is:
\begin{equation}
s = \max_{0 \leq t < B} s^t.
\end{equation}

This protocol measures two core capabilities: \textit{physical inference} (\ie, ``\textit{know what}'': diagnosing hidden parameter changes from feedback) and \textit{mechanism redesign} (\ie, ``\textit{know how}'': revising a structure to function under altered physics). The 20-attempt budget tests whether agents converge through causal reasoning rather than unproductive trial and error. Each base task yields four such pairs ($\tau_0 \to \tau_1, \dots, \tau_0 \to \tau_4$), providing a fine-grained measure across escalating distribution shifts.

\begin{figure*}[t]
  \centering
  \includegraphics[width=0.9\textwidth]{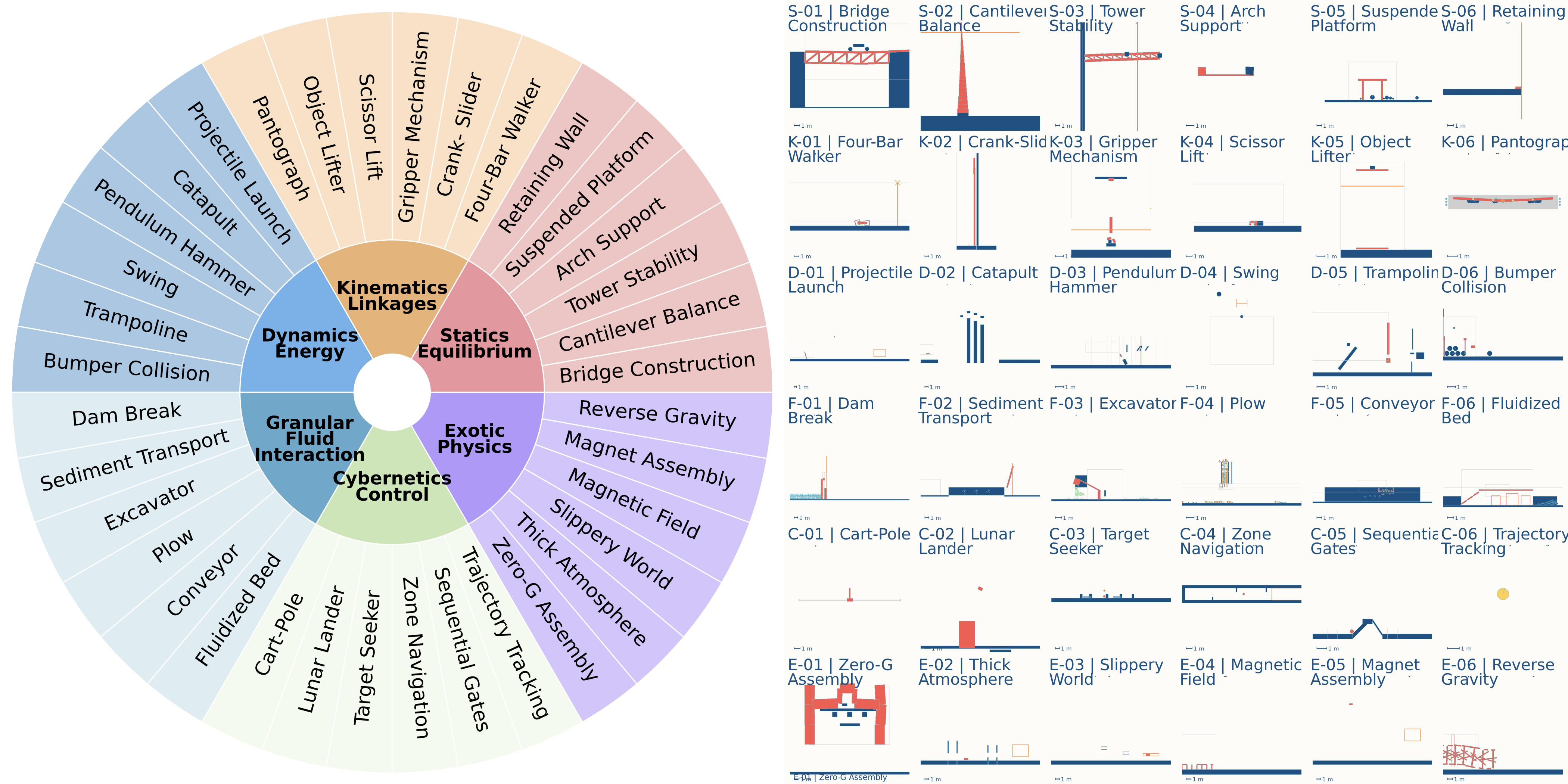}
  \caption{Overview of \textsc{PACE-Bench}. Left: the 36 tasks organized across six physics categories. Right: rendered snapshots of reference solutions succeeding on the source environments.}
  \label{fig:overview}
  \vspace{-0.3cm}
\end{figure*}

\subsection{Dataset Construction}
\label{sec:construction}

We construct each base task by parameterizing $\mathcal{C}$ over $\{\mathcal{E}_0, \dots, \mathcal{E}_4\}$ in a three-phase pipeline (Figure~\ref{fig:construction}).

\noindent\textbf{Phase 1: Task and environment generation.}
We define six physics domains with 6 base tasks per domain. Each base task contributes four source-to-target pairs, yielding $36 \times 4 = 144$ pairs with difficulty increasing from Stage~1 to Stage~4.

\noindent\textbf{Phase 2: Architectural foundation and solution derivation.}
Each task uses a modular architecture of isolated single-responsibility components (Table~\ref{tab:modules}), so parameter changes propagate across all layers. We derive reference solutions for all five environments through iterative refinement with Claude Code\footnote{https://code.claude.com/docs/en/overview}, yielding $\mathcal{X}_k \neq \emptyset$ for all $k$. In sandbox evaluation, the source solution satisfies $x_0 \in \mathcal{X}_0 \setminus \mathcal{X}_k$ at every mutated stage, matching the protocol conditions.

\noindent\textbf{Phase 3: Validation.}
All tasks undergo automated audits for cross-module consistency, variable exposure, and API exclusivity. Prompts disclose explicit constraints and observable variables, omit numeric hidden-physics values such as gravity and friction, and restrict reference solutions to documented primitives (Appendix~\ref{sec:module-audit}). We calibrate difficulty by adjusting mutation parameters (\eg, wider gaps or stricter force limits) until Qwen3-4B fails two independent runs while the task remains solvable (Appendix~\ref{sec:diff-escalation}). We then analyze its logs and add diagnostic measurements to \texttt{feedback.py}, such as force or constraint margins with timestamps and locations. These diagnostics report \textit{what failed} and by \textit{what margin} without recommending a fix. If enriched feedback enables Qwen3-4B to pass, we escalate the difficulty again (Appendix~\ref{sec:feedback-opt}). Finally, two authors with graduate-level physics or engineering training audit all 36 tasks over three passes. Moderate-to-critical issues fall from approximately 86\% of tasks in the first pass to 15--20\% in the second and none in the third (Appendix~\ref{sec:human-verification}).

\subsection{Dataset Overview}

Figure~\ref{fig:overview} presents \textsc{PACE-Bench}. Detailed statistics appear in Table~\ref{tab:dataset-stats} in the Appendix. Averaged across 6 categories, a task prompt contains 1,051 tokens, 7 hard constraints, and 8 primitive APIs, with mutated parameters escalating from 2 at Stage~1 to 10 at Stage~4. In total, the benchmark spans 180 evaluation environments (36 tasks $\times$ 5 environments), 37,860 prompt tokens, 273 hard constraints, 292 primitive APIs, and 945 stage-level parameter mutations. Each pair tests whether an agent can infer hidden physical changes from feedback and revise its design, making \textsc{PACE-Bench} the first benchmark to combine environment variation and physics-grounded self-evolving evaluation.

\begin{table*}[t]
  \centering
  \small
  \begin{tabular}{llcccccc}
  \toprule
  & & \multicolumn{2}{c}{\textbf{Qwen3-4B}} & \multicolumn{2}{c}{\textbf{Qwen3-8B}} & \multicolumn{2}{c}{\textbf{Qwen3-14B}} \\
  \cmidrule(lr){3-4} \cmidrule(lr){5-6} \cmidrule(lr){7-8}
  \textbf{Paradigm} & \textbf{Method} & \textbf{Pass@2} & \textbf{Score@2} & \textbf{Pass@2} & \textbf{Score@2} & \textbf{Pass@2} & \textbf{Score@2} \\
  \midrule
    \multirow{3}{*}{Context} & Vanilla & \cellcolor[HTML]{D0E2F1}11.5\,({\scriptsize $\uparrow$0.0}) & \cellcolor[HTML]{D3E4F2}10.7\,({\scriptsize $\uparrow$0.0}) & \cellcolor[HTML]{C1D8EC}15.6\,({\scriptsize $\uparrow$0.0}) & \cellcolor[HTML]{CFE1F1}12.0\,({\scriptsize $\uparrow$0.0}) & \cellcolor[HTML]{6EA5CD}32.0\,({\scriptsize $\uparrow$0.0}) & \cellcolor[HTML]{8FBADA}25.5\,({\scriptsize $\uparrow$0.0}) \\
    & Reflexion & \cellcolor[HTML]{AECCE5}19.5\,({\scriptsize $\uparrow$8.0}) & \cellcolor[HTML]{C5DBED}14.6\,({\scriptsize $\uparrow$3.9}) & \cellcolor[HTML]{8AB6D8}26.6\,({\scriptsize $\uparrow$11.0}) & \cellcolor[HTML]{B5D1E8}18.1\,({\scriptsize $\uparrow$6.1}) & \cellcolor[HTML]{5D9AC6}35.9\,({\scriptsize $\uparrow$3.9}) & \cellcolor[HTML]{83B2D6}28.0\,({\scriptsize $\uparrow$2.5}) \\
    & Self-Refine & \cellcolor[HTML]{E3EDF7}6.2\,({\scriptsize $\downarrow$5.3}) & \cellcolor[HTML]{E5EFF7}5.5\,({\scriptsize $\downarrow$5.2}) & \cellcolor[HTML]{ECF3F9}3.1\,({\scriptsize $\downarrow$12.5}) & \cellcolor[HTML]{EAF2F9}3.7\,({\scriptsize $\downarrow$8.3}) & \cellcolor[HTML]{E0ECF6}7.1\,({\scriptsize $\downarrow$24.9}) & \cellcolor[HTML]{DEEBF6}8.0\,({\scriptsize $\downarrow$17.5}) \\
  \midrule
    \multirow{3}{*}{Memory} & ACE & \cellcolor[HTML]{C1D9EC}15.5\,({\scriptsize $\uparrow$4.0}) & \cellcolor[HTML]{CADEEF}13.3\,({\scriptsize $\uparrow$2.6}) & \cellcolor[HTML]{AECCE5}19.5\,({\scriptsize $\uparrow$3.9}) & \cellcolor[HTML]{C4DAED}14.9\,({\scriptsize $\uparrow$2.9}) & \cellcolor[HTML]{92BBDC}25.0\,({\scriptsize $\downarrow$7.0}) & \cellcolor[HTML]{ADCCE5}19.6\,({\scriptsize $\downarrow$5.9}) \\
    & ExpeL & \cellcolor[HTML]{CDE0F0}11.5\,({\scriptsize $\uparrow$0.0}) & \cellcolor[HTML]{D4E5F2}10.5\,({\scriptsize $\downarrow$0.2}) & \cellcolor[HTML]{CADEEF}13.2\,({\scriptsize $\downarrow$2.4}) & \cellcolor[HTML]{D6E5F3}10.1\,({\scriptsize $\downarrow$1.9}) & \cellcolor[HTML]{C1D8EC}15.6\,({\scriptsize $\downarrow$16.4}) & \cellcolor[HTML]{CBDFEF}13.0\,({\scriptsize $\downarrow$12.5}) \\
    & ReasoningBank & \cellcolor[HTML]{D2E4F2}10.9\,({\scriptsize $\downarrow$0.6}) & \cellcolor[HTML]{D9E8F4}9.2\,({\scriptsize $\downarrow$1.5}) & \cellcolor[HTML]{B2CFE7}18.8\,({\scriptsize $\uparrow$3.2}) & \cellcolor[HTML]{C7DCEE}14.0\,({\scriptsize $\uparrow$2.0}) & \cellcolor[HTML]{96BEDD}24.2\,({\scriptsize $\downarrow$7.8}) & \cellcolor[HTML]{AECDE5}19.5\,({\scriptsize $\downarrow$6.0}) \\
  \midrule
    \multirow{2}{*}{Search} & ToT & \cellcolor[HTML]{9AC0DE}23.4\,({\scriptsize $\uparrow$11.9}) & \cellcolor[HTML]{B6D2E8}17.9\,({\scriptsize $\uparrow$7.2}) & \cellcolor[HTML]{BAD4EA}17.1\,({\scriptsize $\uparrow$1.5}) & \cellcolor[HTML]{C5DBED}14.5\,({\scriptsize $\uparrow$2.5}) & \cellcolor[HTML]{AACAE4}20.3\,({\scriptsize $\downarrow$11.7}) & \cellcolor[HTML]{BDD6EB}16.5\,({\scriptsize $\downarrow$9.0}) \\
    & CodeEvolve & \cellcolor[HTML]{D3E4F2}10.7\,({\scriptsize $\downarrow$0.8}) & \cellcolor[HTML]{E4EEF7}5.8\,({\scriptsize $\downarrow$4.9}) & \cellcolor[HTML]{E5EFF7}5.3\,({\scriptsize $\downarrow$10.3}) & \cellcolor[HTML]{EBF2F9}3.1\,({\scriptsize $\downarrow$8.9}) & \cellcolor[HTML]{E5EFF7}5.3\,({\scriptsize $\downarrow$26.7}) & \cellcolor[HTML]{EBF2F9}3.3\,({\scriptsize $\downarrow$22.2}) \\
  \midrule
    \multirow{3}{*}{Parameter} & TTT-Discover & \cellcolor[HTML]{E2EDF7}6.5\,({\scriptsize $\downarrow$5.0}) & \cellcolor[HTML]{D1E3F1}11.3\,({\scriptsize $\uparrow$0.6}) & \cellcolor[HTML]{DFEBF6}7.6\,({\scriptsize $\downarrow$8.0}) & \cellcolor[HTML]{CDE0F0}12.3\,({\scriptsize $\uparrow$0.3}) & \cellcolor[HTML]{C6DCEE}14.3\,({\scriptsize $\downarrow$17.7}) & \cellcolor[HTML]{B9D3E9}17.3\,({\scriptsize $\downarrow$8.2}) \\
    & RAGEN & \cellcolor[HTML]{DEEBF6}7.7\,({\scriptsize $\downarrow$3.8}) & \cellcolor[HTML]{CEE0F0}12.2\,({\scriptsize $\uparrow$1.5}) & \cellcolor[HTML]{E0ECF6}7.0\,({\scriptsize $\downarrow$8.6}) & \cellcolor[HTML]{C2D9EC}15.4\,({\scriptsize $\uparrow$3.4}) & \cellcolor[HTML]{C5DBED}14.5\,({\scriptsize $\downarrow$17.5}) & \cellcolor[HTML]{8BB7D9}26.4\,({\scriptsize $\uparrow$0.9}) \\
    & SEAL & \cellcolor[HTML]{D4E5F2}10.4\,({\scriptsize $\downarrow$1.1}) & \cellcolor[HTML]{BBD5EA}17.0\,({\scriptsize $\uparrow$6.3}) & \cellcolor[HTML]{C2D9EC}15.4\,({\scriptsize $\downarrow$0.2}) & \cellcolor[HTML]{B6D1E8}17.9\,({\scriptsize $\uparrow$5.9}) & \cellcolor[HTML]{ACCBE5}20.0\,({\scriptsize $\downarrow$12.0}) & \cellcolor[HTML]{9AC0DE}23.5\,({\scriptsize $\downarrow$2.0}) \\
  \bottomrule
  \end{tabular}
   \caption{Main Results: Pass@2 (\%) and Score@2 by Method and Model. Values in parentheses show $\Delta$ vs.\ Vanilla ($\uparrow$ = improvement, $\downarrow$ = degradation).}
  \label{tab:main_results}
  \vspace{-0.2cm}
\end{table*}

\section{Experiments}

\subsection{Experimental Setup}

\subsubsection{Baseline}

\noindent\textbf{Models.} Because sweeping larger models across all tasks and methods is prohibitively expensive (cost details in Appendix~\ref{sec:cost}), we use two complementary scopes. The full benchmark evaluates Vanilla and all ten self-evolving methods with Qwen3-4B, Qwen3-8B, and Qwen3-14B \cite{yang2025qwen3technicalreport}. On the Statics subset (24 environment pairs), a 20-attempt Vanilla study adds Qwen3-32B, DeepSeek-V4-Pro, and GPT-5.5 to test whether larger models close the performance gap under the full budget. A separate five-attempt Vanilla study broadens frontier-model coverage at lower cost with GPT-5.5 \cite{openai2025gpt5}, Gemini-3.1-Pro \cite{google2025gemini3}, Claude-Opus-4.7 \cite{anthropic2025claude47}, DeepSeek-V4-Pro \cite{deepseek2025v4pro}, Kimi-K2.6 \cite{kimi}, and MiniMax-M2.7 \cite{minimax}. All models use \texttt{thinking} mode.

\noindent\textbf{Methods.} Under the protocol in~\S\ref{sec:eval-protocol}, Vanilla iteratively submits candidate solutions and revises them from diagnostic feedback. Ten self-evolving methods extend this protocol across four paradigms: \textbf{Context-based}: Reflexion \cite{shinn2023reflexion}, Self-Refine \cite{madaan2023self}; \textbf{Memory-augmented}: ACE \cite{zhang2025agentic}, ExpeL \cite{zhao2024expel}, ReasoningBank \cite{ouyang2025reasoningbank}; \textbf{Inference-time Search}: Tree-of-Thoughts (ToT) \cite{yao2023tree}, CodeEvolve \cite{assumpccao2025codeevolve}; and \textbf{Parameter-based}: SEAL \cite{zweiger2026self}, RAGEN \cite{wang2025ragen}, TTT-Discover \cite{yuksekgonul2026learning}. Detailed adaptation notes are in Appendix~\ref{sec:method-adaptation}.

By default, agents see only the Uniform Suffix $\mathcal{U}$ (\S\ref{sec:eval-protocol}), which lists variables that \textit{might} differ across stages without identifying the actual changes. Appendix~\ref{sec:eval-details} provides implementation details, including history management, infrastructure, and costs.

\subsubsection{Evaluation Metrics}
\label{sec:eval-metrics}

We report two primary metrics over two independent runs per environment pair. \textbf{Pass@2} is the fraction of environment pairs where at least one run succeeds. \textbf{Score@2} is the mean of the two run-best scores, with per-attempt scores in $[-100, 100]$. Runs lacking a complete executable solution due to output truncation, parse failure, sandbox crashes, or context overflow are excluded; Appendix~\ref{sec:result-validity} details these exclusions. Appendix~\ref{sec:run-reliability} also reports reliability statistics and the cost rationale for the two-run protocol.

\subsection{Results and Analysis}

We organize the experiments around three progressive questions: (RQ1) how well current self-evolving methods perform under dynamic physics; (RQ2) what makes these tasks hard and how agents fail; and (RQ3) which design choices and interventions improve performance.

\subsubsection{RQ1: How well do current self-evolving methods perform under dynamic physics?}

\begin{tcolorbox}[
    enhanced, breakable, colback=blue!3!white, colframe=blue!15!white,
    boxrule=0.5pt, arc=2mm, drop fuzzy shadow,
    left=12pt, right=12pt, top=8pt, bottom=8pt,
]
\small\faLightbulb\ \textbf{Takeaway 1:} Reflexion leads in Pass@2 and ToT in wall-clock efficiency. Memory-augmented methods anchor search to early designs, while Self-Refine's unverified inner loop compounds errors and adds latency.
\end{tcolorbox}

\begin{table}[t]
  \centering
  \small
  \begin{tabular}{lrr}
    \toprule
    \textbf{Model} & \textbf{Pass@2 (\%)} & \textbf{Score@2} \\
    \midrule
    Qwen3-4B & 8.3 & 12.0 \\
    Qwen3-8B & 33.3 & 21.8 \\
    Qwen3-14B & 37.5 & 27.3 \\
    Qwen3-32B & 37.5 & 28.4 \\
    DeepSeek-V4-Pro & 45.8 & 48.7 \\
    GPT-5.5 & \textbf{66.7} & \textbf{78.1} \\
    \bottomrule
  \end{tabular}
  \caption{Vanilla performance on Statics subset under the full 20-attempt budget.}
  \label{tab:full_budget_models}
  \vspace{-0.2cm}
\end{table}

\begin{table*}[t]
  \centering
  \scriptsize
  \resizebox{\textwidth}{!}{%
  \begin{tabular}{llccccccccccc}
    \toprule
    \textbf{Model} & \textbf{Metric} & \textbf{Vanilla} & \textbf{Reflexion} & \textbf{Self-Refine} & \textbf{ACE} & \textbf{ExpeL} & \makecell{\textbf{Reasoning}\\\textbf{Bank}} & \textbf{ToT} & \makecell{\textbf{Code}\\\textbf{Evolve}} & \textbf{SEAL} & \textbf{RAGEN} & \makecell{\textbf{TTT-}\\\textbf{Discover}} \\
    \midrule
    \multirow{3}{*}{Qwen3-4B}
      & Score & 24.8 & 23.7 & 11.9 & 23.9 & 21.5 & 21.8 & \textbf{27.4} & 13.4 & 23.3 & 13.5 & 14.2 \\
      & Time & 0.9h & 0.8h & 6.0h & 1.3h & 1.4h & 0.8h & \textbf{0.6h} & 1.8h & 2.0h & 5.5h & 5.4h \\
      & S/Hr & 27.5 & 29.6 & 2.0 & 18.4 & 15.4 & 27.2 & \textbf{45.7} & 7.5 & 11.6 & 2.5 & 2.6 \\
    \midrule
    \multirow{3}{*}{Qwen3-8B}
      & Score & 25.5 & 22.1 & 15.0 & 25.7 & 22.6 & 25.0 & 23.2 & 8.9 & \textbf{28.0} & 14.2 & 14.7 \\
      & Time & 1.1h & 1.5h & 5.1h & 1.3h & 1.4h & 1.0h & \textbf{0.5h} & 2.2h & 2.5h & 5.8h & 5.7h \\
      & S/Hr & 23.2 & 14.7 & 2.9 & 19.8 & 16.1 & 25.0 & \textbf{46.4} & 4.1 & 11.2 & 2.4 & 2.6 \\
    \midrule
    \multirow{3}{*}{Qwen3-14B}
      & Score & 25.2 & \textbf{38.1} & 18.0 & 26.3 & 24.8 & 25.5 & 28.9 & 8.9 & 35.0 & 37.1 & 16.4 \\
      & Time & 0.9h & 1.5h & 5.0h & 1.5h & 1.6h & 1.0h & \textbf{0.5h} & 2.4h & 2.8h & 5.7h & 5.6h \\
      & S/Hr & 28.0 & 25.4 & 3.6 & 17.5 & 15.5 & 25.5 & \textbf{57.8} & 3.7 & 12.5 & 6.5 & 2.9 \\
    \bottomrule
  \end{tabular}
  }
  \caption{Cost-normalized results on the Kinematics subset under
  20-attempt budget. Scores are fitted to the nearest $1/24$ increment and shown to one decimal. S/Hr is computed from the fitted scores before display rounding.}
  \label{tab:cost_normalized_category2}
    \vspace{-0.2cm}
\end{table*}

\noindent\textbf{Overall Performance.}
Table~\ref{tab:main_results} reports Pass@2 and Score@2 across methods and models. The strongest configuration, Reflexion + Qwen3-14B, reaches only 35.9\% Pass@2. Gains are larger from 8B to 14B than from 4B to 8B (Figure~\ref{fig:discovery_by_model} in Appendix). To test whether larger models close this gap, we evaluate Vanilla on 24 environment pairs from the Statics subset under the full 20-attempt budget, adding Qwen3-32B, DeepSeek-V4-Pro, and GPT-5.5. Table~\ref{tab:full_budget_models} suggests a within-family plateau: Qwen3-32B does not improve Pass@2 over Qwen3-14B, whereas DeepSeek-V4-Pro and GPT-5.5 perform better. Even GPT-5.5 fails one third of the pairs, leaving the benchmark unsaturated. A complementary five-attempt comparison broadens coverage to six frontier LLMs in Appendix~\ref{sec:model-comparison-details}.

\noindent\textbf{From-Scratch Difficulty.} To test whether the tasks remain non-trivial when agents must construct solutions from scratch, we run Vanilla on all five environments for each task. Each agent receives the task description and primitive APIs, then revises its solution with simulator feedback for up to 20 attempts. Table~\ref{tab:from-scratch-vanilla} reports low Pass@2 across model scales (11.3--18.3\%), showing that direct solution construction remains non-trivial. Table~\ref{tab:from-scratch-environment} shows a much higher Pass@2 on the source environment (32.3\%) than on the target environments (6.4--13.4\%). Thus, our mutations create substantially harder targets, motivating source-to-target adaptation via re-design.

\noindent\textbf{Analysis by Method Paradigm.}
Table~\ref{tab:main_results} reveals clear paradigm-level patterns. \textbf{Context-based.} \textbf{Reflexion} leads overall (35.9\% Pass@2 at 14B, +4--11 points over Vanilla), while \textbf{Self-Refine}, which revises up to five times before sandbox evaluation, never exceeds 7.1\%. This gap suggests that unverified revision compounds errors rather than correcting them. \textbf{Memory-augmented.} \textbf{ACE} and \textbf{ReasoningBank} outperform Vanilla at 8B but underperform at 14B (25.0\%, 24.2\% vs.\ 32.0\%), suggesting retrieved experiences constrain stronger models. \textbf{ExpeL} also trails Vanilla at both scales, possibly because its frozen source-environment memory transfers poorly. \textbf{Inference-time Search.} \textbf{ToT} excels at 4B (23.4\%, 2.0$\times$ Vanilla), but its advantage erodes at larger scales, while \textbf{CodeEvolve} scales negatively (10.7\%$\to$5.3\%). \textbf{Parameter-based.} All three methods underperform Vanilla at 14B. \textbf{SEAL} is strongest (10.4--20.0\%), \textbf{RAGEN}'s gap widens with scale ($-3.8 \to -8.6 \to -17.5$), and \textbf{TTT-Discover} improves at 14B (14.3\%) but remains below Vanilla. Detailed failure analysis via error taxonomy and code similarity is in RQ2 (\S\ref{sec:rq2}).

\noindent\textbf{Cost-Normalized Comparison.}
To account for unequal compute overhead, we compare all eleven methods on the Kinematics subset with Qwen3-4B/8B/14B under the 20-attempt budget. Table~\ref{tab:cost_normalized_category2} reports runtime in hours and Score/Hr, defined as average score divided by runtime. At the paradigm level, inference-time search splits sharply. \textbf{ToT} is the most efficient method at every scale (45.7--57.8 S/Hr), whereas \textbf{CodeEvolve} remains low (3.7--7.5 S/Hr). Context-based methods also diverge. \textbf{Reflexion} combines low overhead with strong efficiency, while \textbf{Self-Refine}'s unverified inner loop is slow and yields only 2.0--3.6 S/Hr. Memory-augmented methods achieve moderate efficiency, and parameter-based methods incur high runtime without a consistent efficiency gain. Thus, \textbf{ToT} remains the efficiency leader even though the highest absolute score varies by scale, suggesting that additional self-evolution compute does not reliably improve cost-normalized performance.

\begin{figure*}[t]
  \centering
  \includegraphics[width=0.9\textwidth]{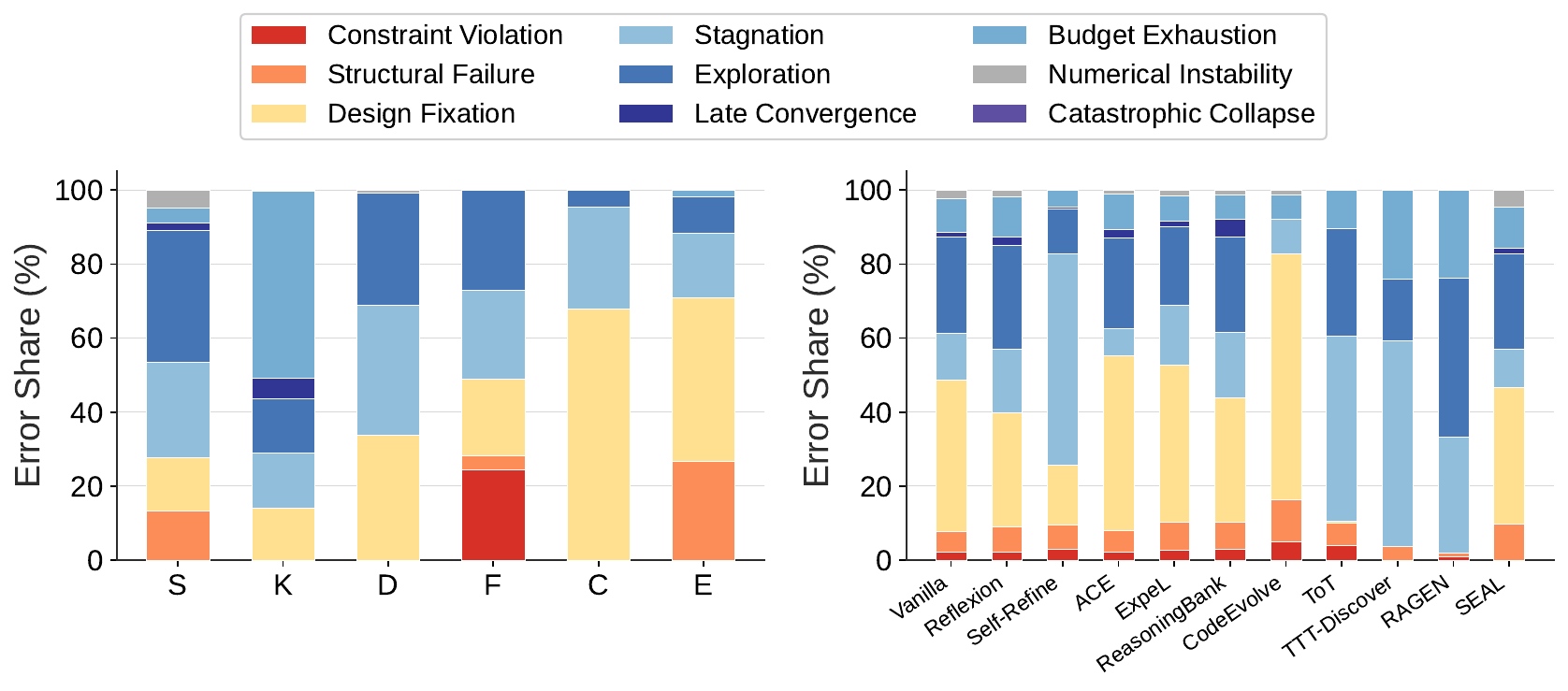}
  \caption{Error type distribution by category (left) and method (right).}
  \label{fig:bar_error_combined}
\end{figure*}

\begin{table*}[t]
    \centering
    \small
    \resizebox{\textwidth}{!}{
    \begin{tabular}{lccccccccc}
        \toprule
         & \multicolumn{3}{c}{\textbf{Qwen3-4B}} & \multicolumn{3}{c}{\textbf{Qwen3-8B}} & \multicolumn{3}{c}{\textbf{Qwen3-14B}} \\
        \cmidrule(lr){2-4} \cmidrule(lr){5-7} \cmidrule(lr){8-10}
        \textbf{Method} & \textbf{Global Sim} & \textbf{Trend (L$-$E)} & \textbf{Radicality} & \textbf{Global Sim} & \textbf{Trend (L$-$E)} & \textbf{Radicality} & \textbf{Global Sim} & \textbf{Trend (L$-$E)} & \textbf{Radicality} \\
        \midrule
        Vanilla & 0.792 & +0.112 & 0.208 & 0.780 & +0.126 & 0.220 & 0.760 & +0.114 & 0.240 \\
        Reflexion & 0.772 & +0.111 & 0.228 & 0.725 & +0.117 & 0.275 & 0.720 & +0.110 & 0.280 \\
        Self-Refine & 0.672 & +0.120 & \textbf{0.328} & 0.598 & +0.122 & \textbf{0.402} & 0.597 & +0.119 & \textbf{0.403} \\
        \cmidrule(lr){1-10}
        ACE & 0.821 & +0.145 & 0.179 & 0.795 & +0.164 & 0.205 & \textbf{0.807} & +0.163 & 0.193 \\
        ExpeL & \textbf{0.840} & +0.132 & 0.160 & \textbf{0.807} & +0.152 & 0.193 & 0.794 & +0.139 & 0.206 \\
        ReasoningBank & 0.760 & \textbf{+0.207} & 0.240 & 0.738 & \textbf{+0.198} & 0.262 & 0.723 & \textbf{+0.233} & 0.277 \\
        \cmidrule(lr){1-10}
        SEAL & 0.755 & +0.097 & 0.245 & 0.805 & +0.117 & 0.195 & 0.726 & +0.096 & 0.274 \\
        \bottomrule
    \end{tabular}
    }
     \caption{Code Similarity by Method and Model (iterations $\geq$ 10). Trend (L$-$E) = mean similarity difference between late and early attempts; positive = increasing self-similarity. Bold denotes the highest value in each column.}
    \label{tab:code_sim_combined}
    \vspace{-0.2cm}
\end{table*}

\subsubsection{RQ2: What makes these tasks hard, and how do agents fail?}
\label{sec:rq2}

\begin{tcolorbox}[
    enhanced, breakable, colback=blue!3!white, colframe=blue!15!white,
    boxrule=0.5pt, arc=2mm, drop fuzzy shadow,
    left=12pt, right=12pt, top=8pt, bottom=8pt,
]
\small\faLightbulb\ \textbf{Takeaway 2:} Self-evolving agents either fixate on early designs or explore without converging. Model scaling appears to improve reasoning more than combinatorial search.
\end{tcolorbox}

\noindent\textbf{Analysis Methods.}
We use two complementary analyses. \textbf{Code similarity analysis.} We compute pairwise Jaccard similarity among generated solutions within each run and report global similarity (mean pairwise similarity), convergence trend (the change from early to late attempts), and radicality ($1 - \text{global similarity}$). Positive convergence trends indicate increasing self-similarity. Table~\ref{tab:code_sim_combined} compares these metrics across methods and model scales, revealing whether agents lock onto a narrow design family or continue making large changes.\footnote{Methods with $<10$ total iterations (multiple attempts per iteration) are excluded for meaningful trend analysis.} \textbf{Error taxonomy.} Every failed run is assigned one of nine mutually exclusive types. Hard failures detected from simulation diagnostics include Catastrophic Collapse, Constraint Violation, Structural Failure, and Numerical Instability. Trajectory-level patterns identified from score progression and code similarity include Design Fixation, Stagnation, Exploration, Late Convergence, and Budget Exhaustion. A priority-ordered pipeline checks hard failures first, then trajectory-level patterns, with the first match determining the type. Details are in Appendix~\ref{sec:error-taxonomy}.

\begin{table*}[t]
  \centering
  \small
  \begin{tabular}{lrrr rrr rrr}
    \toprule
    & \multicolumn{3}{c}{\textbf{Qwen3-4B}} & \multicolumn{3}{c}{\textbf{Qwen3-8B}} & \multicolumn{3}{c}{\textbf{Qwen3-14B}} \\
    \cmidrule(lr){2-4}\cmidrule(lr){5-7}\cmidrule(lr){8-10}
    \textbf{Method} & \textbf{CH} & \textbf{CE} & \textbf{$\Delta$} & \textbf{CH} & \textbf{CE} & \textbf{$\Delta$} & \textbf{CH} & \textbf{CE} & \textbf{$\Delta$} \\
    \midrule
    Vanilla & 1.6 & 10.0 & \textbf{+8.4} & 13.3 & 13.3 & +0.0 & 17.0 & 9.8 & \textbf{$-7.2$} \\
    Reflexion & 9.4 & 12.0 & \textbf{+2.6} & \textbf{17.9} & 13.2 & $-4.7$ & 17.1 & \textbf{14.6} & $-2.5$ \\
    ACE & 6.0 & 12.0 & \textbf{+6.0} & 12.5 & 8.8 & $-3.7$ & 7.3 & 10.4 & +3.1 \\
    \bottomrule
  \end{tabular}
  \caption{Change-Hidden (CH) vs.\ Change-Exposed (CE) Pass@2 (\%) pooled over Statics and Kinematics (approximately 48 environment pairs). CE discloses the changed variables and their exact new values.}
  \label{tab:ce_pooled_categories12}
  \vspace{-0.2cm}
\end{table*}

\noindent\textbf{Analysis by Method.}
Figure~\ref{fig:bar_error_combined} and Table~\ref{tab:code_sim_combined} indicate a split between exploration and exploitation behind the RQ1 performance patterns: conservative methods become trapped in self-similar designs, whereas highly radical methods often change code without making directed progress.

\textbf{Context-based.} \textbf{Vanilla} serves as the baseline: 41.0\% Design Fixation, global similarity 0.777, positive trend (+0.117). The base model converges early to a narrow design region and rarely escapes. \textbf{Self-Refine}'s unverified inner-loop revisions produce chaotic, undirected changes (highest radicality, Stagnation 57.1\%), which may explain why it ranks lowest.

\textbf{Memory-augmented.} \textbf{ACE} cuts Stagnation to 7.5\% (vs.\ Vanilla 12.6\%) but drives Fixation to 47.1\%: score-ranked retrieval repeatedly returns early moderate successes, which are reinserted into later prompts and create a positive-feedback anchoring loop. \textbf{ReasoningBank} has the steepest convergence trend (+0.233) of any method, suggesting progressive anchoring. At 14B, ACE's global similarity peaks at 0.807, indicating that memory may constrain models which reason better from scratch. \textbf{ExpeL}'s Fixation (42.5\%) exceeds Vanilla's while Exploration (21.1\%) is lower, suggesting that frozen source-environment memory narrows rather than guides search. To probe score-retrieval anchoring over time, we include a case study comparing adjacent-attempt similarity for Vanilla and ACE over one 20-attempt trajectory in Appendix~\ref{sec:per-attempt-similarity}.

\textbf{Inference-time Search.} \textbf{ToT} has near-zero Fixation (0.3\%) but 50.2\% Stagnation: breadth-first exploration cannot converge within 20 attempts. \textbf{CodeEvolve} is the inverse, with 66.3\% Fixation (highest), 0.0\% Exploration, 0.835 global similarity, and a flat trend. Its LLM-generated crossover and mutation variants remain too close to selected parent templates, so the population collapses prematurely.

\textbf{Parameter-based.} \textbf{SEAL}'s error profile mirrors Vanilla's (Fixation 36.8\% vs.\ 41.0\%, Stagnation 10.5\% vs.\ 12.6\%), because supervised fine-tuning on its own successes preserves rather than reshapes the base generation distribution. \textbf{RAGEN} has zero Fixation (weight perturbation promotes generation diversity) yet 42.9\% Exploration and 31.4\% Stagnation: the continuous reward is informative but 20 rollouts are too few for RL convergence. \textbf{TTT-Discover} has the highest Stagnation (55.6\%) with zero Fixation because group-based advantage estimation degenerates when most rollouts score identically.

\noindent\textbf{Analysis by Task Category.}
Figure~\ref{fig:bar_error_combined} reports the error taxonomy by category. Failure modes vary sharply across categories, with Design Fixation reaching 58.2\% in \textbf{Control}. Figure~\ref{fig:heatmap_model_category_pass} in Appendix further shows that \textbf{Exotic Physics} is the most solvable category while \textbf{Dynamics} is the hardest across model scales. Task category predicts difficulty far more strongly than model scale: the 4B$\to$14B gain ranges from +16.3 points in \textbf{Exotic Physics} to +1.9 points in \textbf{Kinematics}. Scaling may improve reasoning more than the combinatorial search required by the hardest categories. Detailed per-category error analysis and cross-category comparisons are provided in Appendix~\ref{sec:category-level-analysis}.

\subsubsection{RQ3: What design choices and interventions improve performance?}
\label{sec:rq3}

\begin{tcolorbox}[
    enhanced, breakable, colback=blue!3!white, colframe=blue!15!white,
    boxrule=0.5pt, arc=2mm, drop fuzzy shadow,
    left=12pt, right=12pt, top=8pt, bottom=8pt,
]
\small\faLightbulb\ \textbf{Takeaway 3:} More information is not uniformly helpful: exact parameters do not appear to improve the strongest hidden-setting result, while video feedback tends to help memory-based methods but can hinder context-based methods.
\end{tcolorbox}

\noindent\textbf{Information Asymmetry.}
\label{sec:variable-exposure}
To distinguish ``know what'' from ``know how'', we compare the default \textit{Change-Hidden} (CH) setting with \textit{Change-Exposed} (CE), which reveals the changed variables and their exact values. Table~\ref{tab:ce_pooled_categories12} reports Vanilla, Reflexion, and ACE at all three model scales across Statics and Kinematics. CE improves all 4B configurations (+2.6 to +8.4 points), but at 8B/14B five of six changes remain within $\pm5$ points, while Vanilla-14B drops by 7.2 points. The error taxonomy in Table~\ref{tab:ce_error_shift} shows that CE reduces Design Fixation in five of six ACE/ExpeL settings, suggesting disclosed variables can complement retrieved experiences. Yet ACE still loses 3.7 points at 8B, so this behavioral gain does not consistently translate into Pass@2. Critically, the best CE result (Reflexion-14B, 14.6\%) remains below the best CH result (Reflexion-8B, 17.9\%). Thus, revealing what changed does not raise the performance ceiling, suggesting that mechanism redesign (``\textit{know how}'') is harder than parameter inference (``\textit{know what}'').

\noindent\textbf{VLM Video Feedback.}
\label{sec:vlm-ablation}
To test whether visual evidence resolves failures missed by scalar diagnostics, we augment textual feedback with natural-language failure descriptions generated from videos of candidate executions in the Box2D simulator. An intermediate VLM, Gemma4-26B-A4B \cite{gemmateam2026gemma4technicalreport}, produces these descriptions. We apply this intervention to representative methods on the Statics and Kinematics subsets while keeping the backbone text LLM fixed. Table~\ref{tab:vlm_comparison} in Appendix compares text-only and VLM-augmented feedback. At 4B, additional video feedback helps most methods. At larger scales, it polarizes by paradigm. Context-based methods degrade (Vanilla 14B: $40.9\% \to 18.2\%$, Reflexion: $40.9\% \to 27.3\%$), while memory-augmented methods improve (ACE: $13.6\% \to 22.7\%$, ExpeL: $13.6\% \to 36.4\%$). The error taxonomy across these subsets in Table~\ref{tab:vlm_error_shift} suggests possible mechanisms. For context-based methods, video feedback may scatter focused reasoning, with Exploration rising by over 20 points in several settings. For memory-based methods, fresh visual evidence may counter stale retrieved experiences, with Fixation reductions exceeding 25 points in some settings. Search-based ToT is harmed at 4B and 14B, possibly because video feedback adds redundant variation to already-broad exploration.

\section{Conclusion}

We introduced \textsc{PACE-Bench}, a benchmark of 144 solvable but non-trivial source-to-target environment pairs across six physics domains with systematic hidden environment mutations and simulator-grounded feedback. Within 20 attempts, agents must infer what changed and revise executable code-driven designs to function under target physics. Frontier LLMs remain unsaturated, and most self-evolving methods deliver limited gains over Vanilla or underperform it. Reflexion's explicit simulator-grounded failure analysis improves performance, whereas Self-Refine's unverified inner-loop revisions compound errors. More broadly, agents either fixate on early designs (often in memory-augmented methods) or explore without converging (as in ToT-style inference-time search). Revealing the changed parameters does not improve the performance ceiling, indicating that mechanism redesign (``\textit{know how}'') rather than parameter inference (``\textit{know what}'') is the central bottleneck. Together, these results position \textsc{PACE-Bench} as a testbed for developing this capability.

\section*{Limitations}

\textsc{PACE-Bench} evaluates adaptation in controlled 2D Box2D simulations, so its findings may not transfer to 3D environments or real robots with noisy sensing and actuation. The fixed 20-attempt budget limits the interaction horizon, leaving sustained adaptation untested. Due to computational cost, the frontier-model, parameter-disclosure, and visual-feedback evaluations cover only selected domains and methods. Future work should extend the benchmark to richer simulators and real hardware, longer adaptation horizons, and broader models and methods.

\section*{Ethical Considerations}

\textsc{PACE-Bench} is a research benchmark for evaluating physical intelligence in LLM agents. It does not involve human subjects, personally identifiable information, or deployment in safety-critical systems. All tasks are synthetic and run in simulation with no real-world physical consequences. Because automated mechanical design carries potential dual-use implications, we encourage responsible development. We release the benchmark publicly to promote transparency and reproducibility.

\section*{LLM Usage}                                  During the preparation of this manuscript, we used large language models (LLMs) to assist with language polishing and improving the clarity and readability of the paper. The LLMs were not used to generate research hypotheses, design the methodology, conduct experiments, analyze results, or draw conclusions. All LLM-assisted edits were carefully reviewed and revised by the authors, who take full responsibility for the final content of the manuscript.  



\bibliography{custom}

\clearpage
\appendix

\section{Dataset Construction Details}
\label{sec:construction-details}

\subsection{Module Auditing}
\label{sec:module-audit}

\begin{table}[t]
  \centering
  \small
  \begin{tabular}{lp{0.55\columnwidth}}
    \toprule
    \textbf{Module} & \textbf{Role} \\
    \midrule
    \texttt{environment.py} & Physics and terrain configuration \\
    \texttt{evaluator.py}  & Success criteria and scoring logic \\
    \texttt{feedback.py}   & Diagnostic metrics \\
    \texttt{prompt.py}     & Task context $\mathcal{C}$ \\
    \texttt{stages.py}     & Mutation specs for $\mathcal{E}_{1-4}$ \\
    \texttt{renderer.py}   & Simulation and GIF generation \\
    \texttt{agent.py}      & Reference solutions for $\mathcal{E}_{0-4}$ \\
    \bottomrule
  \end{tabular}
  \caption{Modules of each task in \textsc{PACE-Bench}.}
  \label{tab:modules}
\end{table}

An automated audit script (\texttt{auto\_audit.sh}) systematically checks each task directory against seven rules:
\begin{itemize}[leftmargin=*, nosep, labelsep=5pt]
    \item \textit{Cross-module consistency}: physical parameters, constraints, and success criteria are checked for coherence across \texttt{environment.py}, \texttt{evaluator.py}, \texttt{feedback.py}, \texttt{prompt.py}, \texttt{stages.py}, and \texttt{renderer.py}. See Table~\ref{tab:modules} for the role and responsibility of each module.
    \item \textit{Constraint completeness}: every hardcoded constraint value in \texttt{environment.py} (\eg, mass budgets, force limits) is checked against its numeric value in \texttt{prompt.py}.
    \item \textit{Visible variable synchronization}: every visually observable variable (\eg, gap widths, target positions) is checked against its numeric value in \texttt{prompt.py} and the current environment configuration.
    \item \textit{Invisible variable sweep}: any numeric value of invisible physics variables (\eg, gravity, friction, damping) found in \texttt{prompt.py} is deleted, leaving only qualitative descriptions.
    \item \textit{Mutation synchronization}: when a constraint or visible variable changes across stages in \texttt{stages.py}, the prompt is checked for the new value in the format ``\texttt{new\_value (originally old\_value)}''.
    \item \textit{Uniform Suffix} $\mathcal{U}$ \textit{tone}: the suffix appended to mutated task descriptions is checked for generic warnings without specific numeric values or directions of change.
    \item \textit{Sandbox execution}: every target-environment reference solution must score 100, while the source solution must fail on all four target environments.
\end{itemize}
Violations are reported with file paths and line numbers for immediate correction. The full audit prompt appears in Appendix~\ref{sec:prompts}.

These rules collectively enforce the information asymmetry central to \textsc{PACE-Bench}: constraint and visible variables are always disclosed with precise values, while invisible variables are systematically scrubbed from the prompt and the Uniform Suffix $\mathcal{U}$ provides only generic awareness that shifts may exist, forcing agents to discover the specific physics of each environment through interaction.

\noindent\textbf{Audit Outcomes.}
Table~\ref{tab:human_audit_severity} summarizes the three severity levels used in the human audit and provides a representative example for each. Under this rubric, the first pass identified Lv~2--3 issues in approximately 86\% of tasks. After correction, the second pass found residual issues in approximately 15--20\%. A third pass reran the checklist and found no remaining issues across the 36 tasks. Claude Code assisted construction, reference-solution derivation, feedback design, and difficulty tuning, but it does not judge submitted solutions: hard-coded evaluator logic and Box2D execution determine all scores. Figure~\ref{fig:bar_large_baseline_model} shows that Claude-Opus-4.7 ranks third among the six proprietary models in the constrained-budget study, consistent with the absence of a Claude-specific scoring advantage.

\subsection{Difficulty Escalation}
\label{sec:diff-escalation}

The difficulty escalation pipeline uses Qwen3-4B only as a difficulty check. For each adaptation $\tau_0 \to \tau_k$, the model is evaluated for two independent runs. If it achieves $s(x) = 100$ on any run, the script adjusts the mutation parameters in \texttt{stages.py} to create a harder configuration (\eg, wider gaps, higher gravity, or stricter force limits). After each adjustment, we rerun the target-environment reference solution and revise it if necessary until it scores 100. Only then do we re-evaluate Qwen3-4B. This loop continues until the model fails both runs. Qwen3-4B does not construct the task or determine success criteria, and the target environment is shared unchanged by all evaluated models. The escalation focuses on invisible variables so that difficulty increases cannot be trivially detected from prompt changes alone.

\subsection{Feedback Design}
\label{sec:feedback-opt}

Each task includes a dedicated \texttt{feedback.py} module that translates raw simulation metrics into structured diagnostics. The feedback follows four design principles:
\begin{itemize}[leftmargin=*, nosep, labelsep=5pt]
    \item \textit{Physics-grounded}: reports numerical physical quantities (\eg, peak joint forces and torques, collision impulses, structural deformation, constraint margins) with timestamps and locations, instead of binary pass/fail.
    \item \textit{Dynamic thresholds}: all limit values are read from the current environment configuration rather than hardcoded to maintain accuracy across target environments.
    \item \textit{Diagnostic, not prescriptive}: observations identify \textit{what} failed and \textit{by what margin} (\eg, ``Joint at anchor A broke at step 42 because torque exceeded the limit by 35\%''), without prescribing engineering fixes.
    \item \textit{Phase-aware}: for tasks with distinct simulation phases (\eg, static load, dynamic traversal), metrics are reported per phase to help the agent isolate when and where failures occur.
\end{itemize}

\subsection{Human Verification}
\label{sec:human-verification}

\noindent\textbf{Personnel and Coverage.}
To verify benchmark quality independently of LLM-assisted construction, two authors with graduate-level physics or engineering training divided the six categories, each auditing three categories (18 tasks). All 36 tasks and all LLM-generated modules were manually cross-checked against executable task behavior rather than model judgment.

\noindent\textbf{Protocol.}
For every task, the auditor executed the reference solution on the source and target environments and recorded $s=100$, then ran the source solution and observed failure on all four target environments. The auditor also checked exact numeric consistency across \texttt{environment.py}, \texttt{evaluator.py}, \texttt{feedback.py}, and \texttt{prompt.py}. The same audit scrubbed numeric values of invisible physics variables from prompts, checked the propagation of constraint and visible-variable mutations, and inspected the Uniform Suffix for generic variable awareness only. Together, these sandbox outcomes support the intended solvability and non-triviality conditions without relying on LLM preference.

\begin{table}[t]
  \centering
  \small
  \begin{tabular}{lp{0.63\columnwidth}}
    \toprule
    \textbf{Severity} & \textbf{Definition and example} \\
    \midrule
    Lv~3: Critical & Task unsolvable or evaluator broken (e.g., prompt constraints contradict the evaluator). \\
    Lv~2: Moderate & Task solvable but prompt misleading (e.g., a stale visible-variable value). \\
    Lv~1: Minor & Cosmetic issue (e.g., a renderer color mismatch or unused import). \\
    \bottomrule
  \end{tabular}
  \caption{Severity levels used during human verification.}
  \label{tab:human_audit_severity}
\end{table}

\noindent\textbf{Feedback optimization pipeline.}
We construct feedback modules through a two-phase automated pipeline (\texttt{auto\_feedback.sh}), driven by Claude Code.

\begin{itemize}[leftmargin=*, nosep, labelsep=5pt]
\item{Phase 1: Forensic failure analysis.}
For each task, Qwen3-4B is evaluated on all four target environments. Since Qwen3-4B consistently fails, its execution logs (saved as JSON) provide rich failure samples. An LLM analyzes these logs against six diagnostic dimensions: (1) temporal event chronology, (2) spatial margins to limits, (3) load and stress distribution ranked by severity, (4) energy flow and loss mechanisms (dynamics tasks), (5) constraint satisfaction profile with PASS/FAIL margins for \textit{all} constraints, and (6) numerical health (NaN, Inf, extreme velocities). For each dimension, the analysis identifies missing diagnostic signals, specifically metrics that could have surfaced the root cause. The output is a prioritized list of top 3--5 missing diagnostics, each annotated with the reasoning step it would enable and how it can be computed from existing sandbox state.

\item{Phase 2: Feedback implementation and debloating.}
Guided by the forensic analysis, Claude Code rewrites \texttt{feedback.py} to produce diagnostic reports covering all applicable dimensions for the task type (\eg, temporal chronology and spatial margins for kinematics, and energy chain and loss breakdown for dynamics). The implementation rules are strict: every reported value must be tracked in the metrics dictionary, every threshold must come from \texttt{metrics.get()} rather than a hardcoded literal, and the output must never prescribe engineering fixes. If key simulation quantities are missing from the metrics dict, new keys, tracking variables, or getter methods may be added to \texttt{evaluator.py} and \texttt{environment.py}, but existing defaults, pass/fail logic, and function signatures must remain intact. A subsequent debloating pass trims redundancy: cross-section duplication is eliminated, low-information lists are collapsed to one-line summaries, speculative math is removed, and unchanged sections in multi-moment reports are delta-compressed. The target is a concise, high-signal report where every line carries actionable diagnostic value. The full prompts used for both phases appear in Appendix~\ref{sec:prompts}.

\end{itemize}

\begin{table*}[t]
  \centering\footnotesize
  \resizebox{\textwidth}{!}{
  \begin{tabular}{lrrrrrrr}
    \toprule
    \textbf{Category} & \textbf{\#Prompt Tokens} & \textbf{\#Constraints} & \textbf{\#Prim. APIs} & \textbf{Stage 1} & \textbf{Stage 2} & \textbf{Stage 3} & \textbf{Stage 4} \\
    \midrule
    Statics \& Equilibrium & 935 & 8 & 3 & 4 & 4 & 11 & 11 \\
    Kinematics \& Linkages & 835 & 7 & 9 & 3 & 4 & 7 & 9 \\
    Dynamics \& Energy & 894 & 12 & 9 & 1 & 2 & 5 & 8 \\
    Granular \& Fluid Interaction & 1,136 & 9 & 10 & 2 & 2 & 13 & 12 \\
    Cybernetics \& Control & 1,730 & 5 & 9 & 2 & 4 & 12 & 12 \\
    Exotic Physics & 778 & 1 & 6 & 1 & 4 & 6 & 8 \\
    \midrule
    \textbf{Avg.} & \textbf{1,051} & \textbf{7} & \textbf{8} & \textbf{2} & \textbf{3} & \textbf{9} & \textbf{10} \\
    \textbf{Total} & \textbf{37,860} & \textbf{273} & \textbf{292} & \textbf{94} & \textbf{139} & \textbf{340} & \textbf{372} \\
    \bottomrule
  \end{tabular}
  }
  \caption{Dataset statistics by category (averaged over 6 tasks each). Stages 1--4 report average mutated physical parameters per stage, indicating progressive escalation in difficulty.}
  \label{tab:dataset-stats}
\end{table*}

\section{Error Taxonomy Judging Criteria}
\label{sec:error-taxonomy}

Each failed run is assigned a single error type via a priority-ordered classifier (first matching condition wins). The priority order is: hard failures first (Catastrophic Collapse, Constraint Violation, Structural Failure, Numerical Instability), then trajectory-level patterns (Design Fixation, Late Convergence, Stagnation, Exploration), with Budget Exhaustion as fallback for runs that make progress but exhaust all attempts.

\begin{itemize}[leftmargin=*, nosep, labelsep=5pt]
\item \textbf{Catastrophic Collapse}: best score $\leq -60$.
\item \textbf{Constraint Violation}: any hard constraint flag set in the best attempt.
\item \textbf{Structural Failure}: joint or beam breakage detected.
\item \textbf{Numerical Instability}: NaN, Inf, or solver divergence with best score $<$ 0.
\item \textbf{Design Fixation}: avg code similarity $>$ 0.85 over last 3 iterations, score trend $<$ 3, best score $<$ 50.
\item \textbf{Late Convergence}: best in second half exceeds best in first half by $>$ 2 points, and $0 <$ best score $< 100$.
\item \textbf{Stagnation}: never achieves positive score, flat trajectory, $\leq$1 unique failure type.
\item \textbf{Exploration}: never achieves positive score, $\geq$3 unique failure types.
\item \textbf{Budget Exhaustion}: $0 <$ best score $< 100$, all 20 attempts used, no other type applies.
\end{itemize}

The main-text analysis in Table~\ref{tab:code_sim_combined} reports the cross-model code-similarity evidence used to characterize these trajectory-level failure patterns.

\section{Method Adaptation Details}
\label{sec:method-adaptation}

Each self-evolving method was originally designed for text, code, or discrete action domains. We adapt them to the physics simulation setting of \textsc{PACE-Bench}. All methods share the same base protocol: receive Box2D diagnostic feedback after each attempt, and produce a revised solution (Python program) within a 20-attempt budget. All auxiliary LLM calls (for reflection, memory induction, or rule extraction) use the same backbone solver model, preserving full autonomy.

\noindent\textbf{Baseline.} The Vanilla iterative refinement loop generates code via the LLM, executes it in the Box2D sandbox, appends structured feedback to the revision prompt, and repeats. No extra memory, agents, or weight updates.

\noindent\textbf{Reflexion \cite{shinn2023reflexion}.} After each failed attempt, the backbone model produces a 3--8 sentence diagnosis with a high-level fix plan. Up to 3 reflections are accumulated (FIFO) and injected before the task description in subsequent revision prompts.

\noindent\textbf{Self-Refine \cite{madaan2023self}.} The model self-critiques and self-corrects in an inner loop (max 5 steps) without external verification. The outer verifier runs once after the inner loop. Contradiction detection prevents premature stopping when the model declares correctness but simultaneously outputs new code.

\noindent\textbf{ACE \cite{zhang2025agentic}.} Maintains a structured \textit{playbook} (strategies, code snippets, common mistakes). A Reflector analyzes each attempt, while a Curator merges and prunes the playbook. Both use the backbone model.

\noindent\textbf{ExpeL \cite{zhao2024expel}.} Extracts distilled \textit{rules} from rollout trajectories via LLM critique, retrieved by embedding similarity (Sup-SimCSE). Operates in pair-based mode: rules are learned from the source-environment rollout and injected into the prompt (never raw code, since environment parameters differ across mutations).

\noindent\textbf{ReasoningBank \cite{ouyang2025reasoningbank}.} Induces structured memory items from each attempt via the backbone model. Supports parallel \textit{MaTTS}: $k$ candidates are generated and evaluated in the sandbox, then a contrast-and-distill call extracts what distinguished successes from failures. Success is judged by verifier score $\geq 99$. Memory persists for cross-mutation transfer. Retrieval uses instruction-aware embedding similarity ($k = 5$).

\noindent\textbf{Tree-of-Thoughts \cite{yao2023tree}.} Beam search over code revisions: retain top-$b$ states by verifier score, generate $n$ candidates from each, evaluate all, keep top $b$. Default $b = 3$, $n = 2$. A budget formula matches the attempt budget: $(b \times n + 1) \times \text{rounds} \leq \text{max\_iterations}$.

\noindent\textbf{CodeEvolve \cite{assumpccao2025codeevolve}.} This population-based method uses the LLM as crossover and mutation operator to create variant physics programs, while the Box2D verifier score is fitness for selection. The population improves over generations, with population size 8 and at most 20 total LLM calls. The official CodeEvolve CLI runs as a subprocess with one island per GPU worker.

\noindent\textbf{SEAL \cite{zweiger2026self}.} This supervised test-time method resets LoRA at every iteration, retrains from scratch on all accumulated (prompt, code) pairs with positive Box2D scores, and then generates the next solution. It uses 20 iterations. Training reuses prior verifications and does not add evaluation attempts.

\noindent\textbf{RAGEN \cite{wang2025ragen}.} This online RL method treats code revision as a multi-turn episode (generate $\rightarrow$ execute $\rightarrow$ revise), computes GRPO advantages per turn, and updates LoRA via PPO-clip, thereby learning the refinement trajectory rather than only final outputs. Five iterations $\times$ two episodes $\times$ two turns match the 20-attempt budget. LoRA rank is 64, with learning rate $10^{-5}$ and two PPO epochs.

\noindent\textbf{TTT-Discover \cite{yuksekgonul2026learning}.} This test-time RL method generates independent candidates, computes leave-one-out entropic advantages, and updates LoRA with an importance-sampling loss. Identical candidate scores trigger feedback-driven expansion to break the deadlock. Five iterations $\times$ four rollouts match the 20-attempt budget. After iteration 1, revision prompts condition on prior feedback. The group size is 4, with 50 training epochs, learning rate $4\times10^{-5}$, and LoRA rank 32.

\begin{table*}[t]
  \centering
  \small
  \begin{tabular}{p{0.42\textwidth}l}
    \toprule
    \textbf{Full-benchmark statistic} & \textbf{Value} \\
    \midrule
    Pass/fail agreement between runs & 86.6\% \\
    Median absolute score difference & 0.0 \\
    Environment pairs with score difference $\leq 5$ & 82.0\% \\
    Spearman $\rho$, run 1 vs.\ run 2 (4B / 8B / 14B) & 0.667 / 0.881 / 0.976 \\
    \midrule
    \textbf{Kinematics: two-run vs.\ three-run statistic} & \textbf{Value} \\
    \midrule
    Pair-level pass agreement & 94.9\% \\
    Pearson $r$, pair-level mean scores & 0.940 \\
    Mean absolute error, pair-level mean scores & 5.44 \\
    Spearman $\rho$, method ranks (4B / 8B / 14B) & 1.000 / 0.700 / 0.900 \\
    \bottomrule
  \end{tabular}
  \caption{Run-reliability analysis. The full-benchmark panel compares the two independent runs; the Kinematics panel compares statistics computed from two runs with those computed after adding a third run for five representative methods.}
  \label{tab:run_reliability}
\end{table*}

\begin{figure}[t]
  \centering
  \includegraphics[width=\columnwidth]{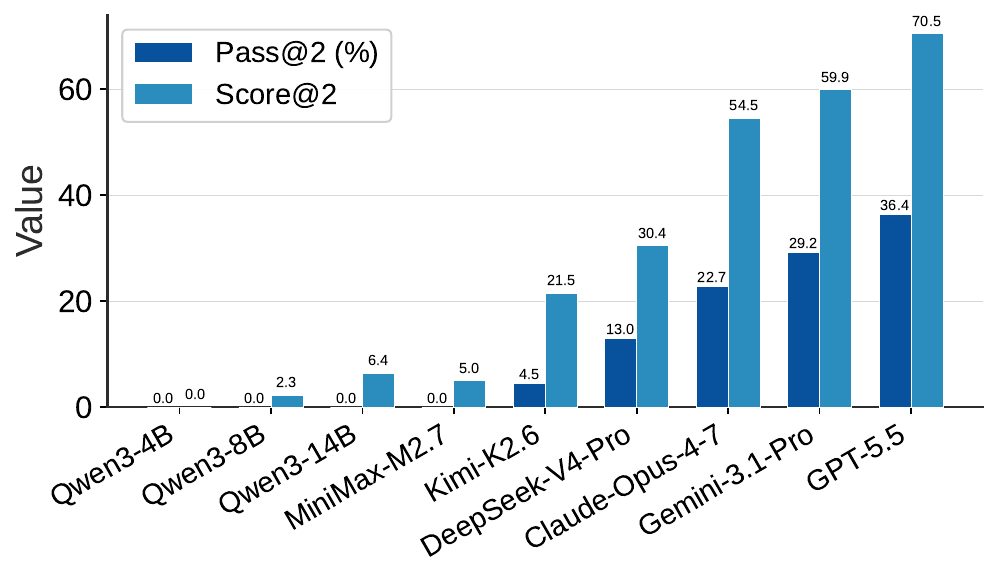}
  \caption{Pass@2 and Score@2 by model on Statics, Vanilla method, five-attempt budget.}
  \label{fig:bar_large_baseline_model}
\end{figure}

\begin{table*}[t]
    \centering
    \resizebox{\textwidth}{!}{
    \begin{tabular}{lcccccc}
        \toprule
        \textbf{Method} & \multicolumn{2}{c}{\textbf{Qwen3-4B}} & \multicolumn{2}{c}{\textbf{Qwen3-8B}} & \multicolumn{2}{c}{\textbf{Qwen3-14B}} \\
        \cmidrule(lr){2-3} \cmidrule(lr){4-5} \cmidrule(lr){6-7}
         & \textbf{Pass@2 w/o VLM} & \textbf{Pass@2 w/ VLM} & \textbf{Pass@2 w/o VLM} & \textbf{Pass@2 w/ VLM} & \textbf{Pass@2 w/o VLM} & \textbf{Pass@2 w/ VLM} \\
        \midrule
        Vanilla & 0.0 & 18.2\,({\scriptsize \textcolor{green!50!black}{+18.2}}) & 24.4 & 17.1\,({\scriptsize \textcolor{red!50!black}{-7.3}}) & 40.9 & 18.2\,({\scriptsize \textcolor{red!50!black}{-22.7}}) \\
        Reflexion & 9.1 & 22.7\,({\scriptsize \textcolor{green!50!black}{+13.6}}) & 42.9 & 23.8\,({\scriptsize \textcolor{red!50!black}{-19.1}}) & 40.9 & 27.3\,({\scriptsize \textcolor{red!50!black}{-13.6}}) \\
        ACE & 4.5 & 13.6\,({\scriptsize \textcolor{green!50!black}{+9.1}}) & 18.2 & 31.8\,({\scriptsize \textcolor{green!50!black}{+13.6}}) & 13.6 & 22.7\,({\scriptsize \textcolor{green!50!black}{+9.1}}) \\
        ExpeL & 13.6 & 9.1\,({\scriptsize \textcolor{red!50!black}{-4.5}}) & 14.3 & 42.9\,({\scriptsize \textcolor{green!50!black}{+28.6}}) & 13.6 & 36.4\,({\scriptsize \textcolor{green!50!black}{+22.8}}) \\
        ToT & 13.6 & 9.1\,({\scriptsize \textcolor{red!50!black}{-4.5}}) & 9.1 & 18.2\,({\scriptsize \textcolor{green!50!black}{+9.1}}) & 4.5 & 0.0\,({\scriptsize \textcolor{red!50!black}{-4.5}}) \\
        \bottomrule
    \end{tabular}
    }
    \caption{VLM Video Feedback Impact on Statics and Kinematics.}
    \label{tab:vlm_comparison}
\end{table*}

\begin{table*}
    \centering
    \resizebox{\textwidth}{!}{
    \begin{tabular}{llccccccccc}
        \toprule
        \multirow{2}{*}{\textbf{Model}} & \multirow{2}{*}{\textbf{Method}} & \textbf{Catastrophic} & \textbf{Structural} & \textbf{Constraint} & \textbf{Numerical} & \textbf{Design} & \multirow{2}{*}{\textbf{Stagnation}} & \multirow{2}{*}{\textbf{Exploration}} & \textbf{Late} & \textbf{Budget} \\
         &  & \textbf{Collapse} & \textbf{Failure} & \textbf{Violation} & \textbf{Instability} & \textbf{Fixation} &  &  & \textbf{Convergence} & \textbf{Exhaustion} \\
        \midrule
        Qwen3-4B & Vanilla & 0.0 & \textcolor{red!60!black}{+3.0} & 0.0 & \textcolor{red!60!black}{+3.0} & \textcolor{red!60!black}{+5.1} & \textcolor{green!50!black}{-8.1} & \textcolor{red!60!black}{+6.1} & \textcolor{green!50!black}{-4.5} & \textcolor{green!50!black}{-4.5} \\
         & Reflexion & 0.0 & \textcolor{red!60!black}{+1.8} & 0.0 & \textcolor{green!50!black}{-9.1} & \textcolor{green!50!black}{-4.1} & \textcolor{green!50!black}{-4.1} & \textcolor{red!60!black}{+24.7} & \textcolor{red!60!black}{+5.9} & \textcolor{green!50!black}{-15.0} \\
         & ACE & 0.0 & \textcolor{red!60!black}{+1.5} & 0.0 & \textcolor{red!60!black}{+11.0} & \textcolor{green!50!black}{-12.3} & \textcolor{red!60!black}{+15.8} & \textcolor{green!50!black}{-11.3} & 0.0 & \textcolor{green!50!black}{-4.8} \\
         & ExpeL & 0.0 & \textcolor{red!60!black}{+4.2} & 0.0 & \textcolor{red!60!black}{+9.7} & \textcolor{green!50!black}{-26.6} & \textcolor{red!60!black}{+4.2} & \textcolor{red!60!black}{+3.4} & \textcolor{red!60!black}{+5.0} & 0.0 \\
         & ToT & 0.0 & \textcolor{red!60!black}{+4.5} & 0.0 & 0.0 & 0.0 & \textcolor{red!60!black}{+18.9} & \textcolor{green!50!black}{-23.2} & 0.0 & 0.0 \\
        \cmidrule(lr){1-11}
        Qwen3-8B & Vanilla & 0.0 & \textcolor{red!60!black}{+2.4} & 0.0 & \textcolor{green!50!black}{-0.6} & \textcolor{green!50!black}{-8.2} & \textcolor{red!60!black}{+2.7} & \textcolor{red!60!black}{+3.0} & \textcolor{red!60!black}{+2.9} & \textcolor{green!50!black}{-2.3} \\
         & Reflexion & 0.0 & \textcolor{green!50!black}{-10.4} & 0.0 & \textcolor{red!60!black}{+2.1} & 0.0 & \textcolor{green!50!black}{-8.3} & \textcolor{red!60!black}{+25.0} & 0.0 & \textcolor{green!50!black}{-8.3} \\
         & ACE & 0.0 & \textcolor{red!60!black}{+8.9} & 0.0 & 0.0 & \textcolor{green!50!black}{-31.1} & \textcolor{green!50!black}{-5.6} & \textcolor{red!60!black}{+33.3} & 0.0 & \textcolor{green!50!black}{-5.6} \\
         & ExpeL & 0.0 & \textcolor{red!60!black}{+5.6} & 0.0 & \textcolor{red!60!black}{+2.8} & \textcolor{green!50!black}{-2.8} & \textcolor{green!50!black}{-8.3} & \textcolor{red!60!black}{+5.6} & \textcolor{red!60!black}{+2.8} & \textcolor{green!50!black}{-5.6} \\
         & ToT & 0.0 & \textcolor{red!60!black}{+6.7} & 0.0 & 0.0 & \textcolor{red!60!black}{+5.6} & \textcolor{green!50!black}{-11.7} & \textcolor{green!50!black}{-6.7} & 0.0 & \textcolor{red!60!black}{+6.1} \\
        \cmidrule(lr){1-11}
        Qwen3-14B & Vanilla & 0.0 & \textcolor{red!60!black}{+1.3} & 0.0 & \textcolor{red!60!black}{+1.3} & \textcolor{green!50!black}{-2.1} & \textcolor{green!50!black}{-15.4} & \textcolor{red!60!black}{+22.6} & \textcolor{green!50!black}{-7.7} & 0.0 \\
         & Reflexion & 0.0 & \textcolor{green!50!black}{-10.6} & 0.0 & 0.0 & \textcolor{red!60!black}{+11.1} & \textcolor{green!50!black}{-7.7} & \textcolor{red!60!black}{+10.1} & \textcolor{green!50!black}{-1.4} & \textcolor{green!50!black}{-1.4} \\
         & ACE & 0.0 & \textcolor{red!60!black}{+6.5} & 0.0 & \textcolor{green!50!black}{-10.5} & \textcolor{red!60!black}{+13.6} & \textcolor{green!50!black}{-3.4} & \textcolor{green!50!black}{-6.8} & \textcolor{green!50!black}{-5.3} & \textcolor{red!60!black}{+5.9} \\
         & ExpeL & 0.0 & \textcolor{green!50!black}{-1.5} & 0.0 & \textcolor{red!60!black}{+3.8} & \textcolor{green!50!black}{-5.3} & \textcolor{red!60!black}{+1.9} & \textcolor{red!60!black}{+4.5} & \textcolor{green!50!black}{-5.3} & \textcolor{red!60!black}{+1.9} \\
         & ToT & 0.0 & \textcolor{green!50!black}{-5.4} & 0.0 & 0.0 & 0.0 & \textcolor{green!50!black}{-6.3} & \textcolor{red!60!black}{+7.4} & 0.0 & \textcolor{red!60!black}{+4.3} \\
        \bottomrule
    \end{tabular}
    }
    \caption{VLM Effect on Error Taxonomy: $\Delta$ error share when VLM video feedback is added. Positive = VLM increases this error type; negative = VLM reduces it.}
    \label{tab:vlm_error_shift}
\end{table*}

\begin{table*}
    \centering
    \resizebox{\textwidth}{!}{
    \begin{tabular}{llccccccccc}
        \toprule
        & & \textbf{Catastrophic} & \textbf{Structural} & \textbf{Constraint} & \textbf{Numerical} & \textbf{Design} &  &  & \textbf{Late} & \textbf{Budget} \\
        \textbf{Model} & \textbf{Method} & \textbf{Collapse} & \textbf{Failure} & \textbf{Violation} & \textbf{Instability} & \textbf{Fixation} & \textbf{Stagnation} & \textbf{Exploration} & \textbf{Convergence} & \textbf{Exhaustion} \\
        \midrule
        Qwen3-4B & Vanilla & 0.0 & \textcolor{red!60!black}{+3.0} & 0.0 & \textcolor{green!50!black}{-2.5} & \textcolor{green!50!black}{-11.6} & \textcolor{green!50!black}{-2.5} & \textcolor{red!60!black}{+22.7} & \textcolor{green!50!black}{-4.5} & \textcolor{green!50!black}{-4.5} \\
         & Reflexion & 0.0 & \textcolor{red!60!black}{+10.0} & 0.0 & \textcolor{red!60!black}{+5.0} & \textcolor{red!60!black}{+3.3} & \textcolor{red!60!black}{+3.3} & \textcolor{green!50!black}{-13.3} & 0.0 & \textcolor{green!50!black}{-8.3} \\
         & ACE & 0.0 & \textcolor{red!60!black}{+2.4} & 0.0 & \textcolor{red!60!black}{+6.3} & \textcolor{green!50!black}{-11.1} & \textcolor{red!60!black}{+5.6} & \textcolor{green!50!black}{-4.0} & \textcolor{red!60!black}{+5.6} & \textcolor{green!50!black}{-4.8} \\
         & ExpeL & 0.0 & \textcolor{green!50!black}{-0.8} & 0.0 & \textcolor{red!60!black}{+4.7} & \textcolor{green!50!black}{-6.6} & \textcolor{green!50!black}{-0.8} & \textcolor{red!60!black}{+3.4} & 0.0 & 0.0 \\
         & ToT & 0.0 & \textcolor{red!60!black}{+5.3} & 0.0 & \textcolor{red!60!black}{+5.3} & 0.0 & \textcolor{red!60!black}{+5.3} & \textcolor{green!50!black}{-15.8} & 0.0 & 0.0 \\
        \cmidrule(lr){1-11}
        Qwen3-8B & Vanilla & 0.0 & \textcolor{green!50!black}{-1.0} & 0.0 & \textcolor{green!50!black}{-14.3} & \textcolor{green!50!black}{-1.4} & \textcolor{red!60!black}{+6.7} & \textcolor{red!60!black}{+17.1} & 0.0 & \textcolor{green!50!black}{-7.1} \\
         & Reflexion & 0.0 & \textcolor{green!50!black}{-4.9} & 0.0 & \textcolor{green!50!black}{-15.4} & \textcolor{red!60!black}{+15.8} & \textcolor{green!50!black}{-7.7} & \textcolor{red!60!black}{+14.6} & \textcolor{red!60!black}{+5.3} & \textcolor{green!50!black}{-7.7} \\
         & ACE & 0.0 & 0.0 & 0.0 & 0.0 & \textcolor{green!50!black}{-16.7} & \textcolor{green!50!black}{-5.6} & \textcolor{red!60!black}{+16.7} & 0.0 & \textcolor{red!60!black}{+5.6} \\
         & ExpeL & 0.0 & \textcolor{red!60!black}{+3.0} & 0.0 & \textcolor{red!60!black}{+7.2} & \textcolor{green!50!black}{-10.5} & \textcolor{green!50!black}{-3.3} & \textcolor{red!60!black}{+1.6} & \textcolor{red!60!black}{+7.2} & \textcolor{green!50!black}{-5.3} \\
         & ToT & 0.0 & \textcolor{red!60!black}{+1.1} & 0.0 & 0.0 & 0.0 & \textcolor{green!50!black}{-0.6} & \textcolor{green!50!black}{-1.1} & 0.0 & \textcolor{red!60!black}{+0.6} \\
        \cmidrule(lr){1-11}
        Qwen3-14B & Vanilla & 0.0 & \textcolor{red!60!black}{+2.3} & 0.0 & \textcolor{red!60!black}{+2.3} & \textcolor{red!60!black}{+4.1} & \textcolor{green!50!black}{-9.5} & \textcolor{red!60!black}{+8.6} & \textcolor{green!50!black}{-7.7} & 0.0 \\
         & Reflexion & 0.0 & 0.0 & 0.0 & \textcolor{red!60!black}{+23.1} & 0.0 & 0.0 & \textcolor{green!50!black}{-7.7} & \textcolor{green!50!black}{-7.7} & \textcolor{green!50!black}{-7.7} \\
         & ACE & 0.0 & \textcolor{red!60!black}{+6.5} & 0.0 & \textcolor{green!50!black}{-4.6} & \textcolor{green!50!black}{-4.0} & \textcolor{green!50!black}{-15.2} & \textcolor{red!60!black}{+22.6} & \textcolor{green!50!black}{-5.3} & 0.0 \\
         & ExpeL & 0.0 & 0.0 & 0.0 & 0.0 & \textcolor{red!60!black}{+5.3} & \textcolor{red!60!black}{+5.3} & \textcolor{green!50!black}{-15.8} & 0.0 & \textcolor{red!60!black}{+5.3} \\
         & ToT & 0.0 & \textcolor{green!50!black}{-4.0} & 0.0 & 0.0 & 0.0 & \textcolor{green!50!black}{-8.1} & \textcolor{red!60!black}{+6.9} & 0.0 & \textcolor{red!60!black}{+5.2} \\
        \bottomrule
    \end{tabular}
    }
    \caption{CE Effect on Error Taxonomy: $\Delta$ error share when mutated variables are disclosed (CE minus CH). Positive = CE increases this error type; negative = CE reduces it. Statics only.}
    \label{tab:ce_error_shift}
\end{table*}

\begin{figure}[t]
  \centering
  \includegraphics[width=\columnwidth]{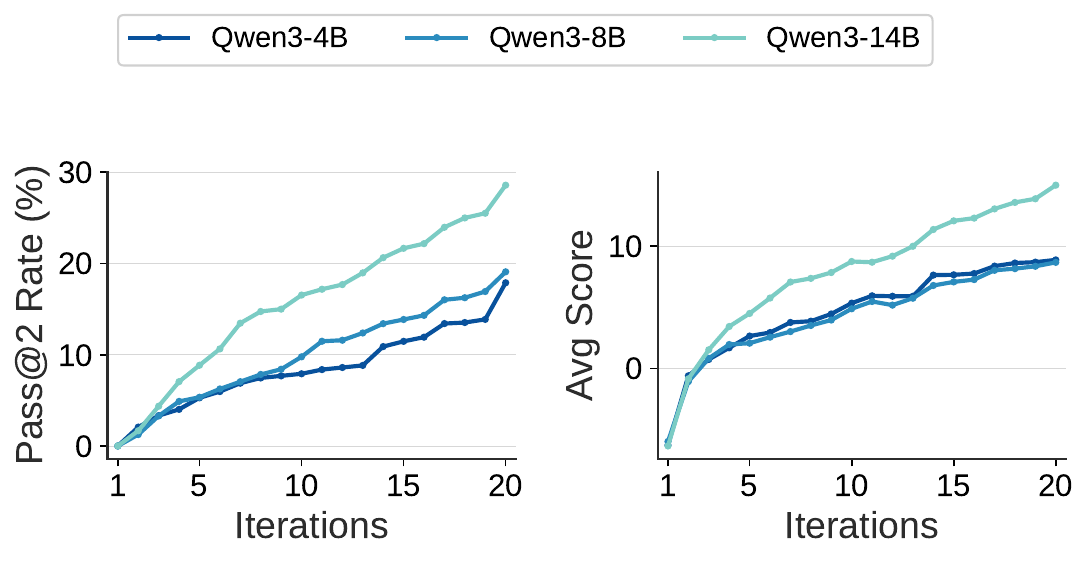}
  \caption{Pass@2@$k$ and Score@2@$k$ by model (all methods, all categories).}
  \label{fig:discovery_by_model}
\end{figure}

\begin{table*}[t]
  \centering
  \scriptsize
  \resizebox{\textwidth}{!}{%
  \begin{tabular}{lrrrrrrrrrr}
    \toprule
    \textbf{Method} & \textbf{1} & \textbf{2} & \textbf{3} & \textbf{4} & \textbf{5} & \textbf{6} & \textbf{7} & \textbf{8} & \textbf{9} & \textbf{10} \\
    \midrule
    Vanilla & 0.255 & 0.468 & 0.536 & 0.843 & 0.518 & 0.493 & 0.778 & 0.793 & 0.553 & 0.943 \\
    ACE & 0.185 & 0.870 & 0.842 & 0.762 & 0.772 & 0.895 & 0.814 & \textbf{1.000} & 0.670 & 0.828 \\
    \bottomrule
  \end{tabular}%
  }

  \vspace{2pt}
  \resizebox{\textwidth}{!}{%
  \begin{tabular}{lrrrrrrrrrr}
    \toprule
    \textbf{Method} & \textbf{11} & \textbf{12} & \textbf{13} & \textbf{14} & \textbf{15} & \textbf{16} & \textbf{17} & \textbf{18} & \textbf{19} & \textbf{20} \\
    \midrule
    Vanilla & 0.524 & \textbf{0.963} & 0.887 & 0.767 & 0.564 & 0.895 & 0.770 & 0.500 & 0.886 & 0.723 \\
    ACE & 0.497 & 0.734 & 0.375 & 0.635 & 0.965 & 0.965 & 0.908 & 0.877 & 0.847 & \textbf{1.000} \\
    \bottomrule
  \end{tabular}%
  }
  \caption{Adjacent-attempt Jaccard code similarity for one Qwen3-14B Stage~4 case study. Each entry compares solutions on consecutive transitions.}
  \label{tab:per_attempt_k06}
\end{table*}

\section{Supplementary Experiment Results}
\label{sec:eval-details}

\subsection{Implementation}
\label{sec:implementation}

\noindent\textbf{Hardware.} Local-model inference runs on 25 NVIDIA A100-80GB GPUs. Qwen3-4B, -8B, -14B, and -32B are served through vLLM with tensor parallelism across 2--4 GPUs per instance. GPT-5.5, Gemini-3.1-Pro, Claude-Opus-4.7, DeepSeek-V4-Pro, Kimi-K2.6, and MiniMax-M2.7 are accessed through their respective cloud APIs.

\noindent\textbf{Hyperparameters.} All methods use temperature $T = 0.7$, top-$p = 0.95$, and a maximum of 65,536 output tokens. The Box2D sandbox runs at 60 FPS (PPM = 40 pixels/meter) with up to 10,000 simulation steps per attempt. Each environment pair ($\tau_0 \to \tau_k$) is evaluated with two independent runs. Pass@2 records whether either run passes, while Score@2 averages the best score obtained within each run.

\noindent\textbf{History Management.} Under the 20-attempt budget, naive retention of the full interaction history $(x^i, s^i, d^i)_{i=0}^{t-1}$ rapidly exceeds LLM context limits (especially for verbose diagnostic feedback and long code outputs). We adopt a \textit{Previous-One + Best} truncation strategy: at each revision step, the agent receives the most recent attempt's full details plus the highest-scoring attempt across the entire history, while intermediate attempts are excluded. This retains both local momentum (the latest result) and global signal (the best result) without context overflow. Preliminary experiments suggest that this scheme preserves over 95\% of the final pass rate compared to the full-history oracle.

\noindent\textbf{Infrastructure.} The evaluation pipeline supports task selection, local/vLLM/API backends, and parallel assignment across methods and tasks. The verifier then enforces API exclusivity, executes candidates in an isolated Box2D sandbox, and logs structured JSON diagnostics.

\subsection{Cost}
\label{sec:cost}

The full local-model sweep evaluates Qwen3-4B, -8B, and -14B on 36 tasks $\times$ 4 environment pairs $\times$ 11 methods $\times$ 3 model sizes $\times$ 2 runs, totaling approximately 9,500 trajectories and up to 190K LLM calls over 4,200 A100 GPU-hours. One additional run at the same scope would require approximately 2,100 A100 GPU-hours. The constrained API study evaluates six proprietary models with Vanilla on 6 tasks $\times$ 4 environment pairs $\times$ 2 runs $\times$ 5 attempts, totaling 1,440 calls and approximately \$2,000. An extra run would add approximately \$1,000. At the observed average API rate, extending these models to all 36 tasks, 4 environment pairs, 11 methods, and 20 attempts would cost approximately \$264,000 per run.

\subsection{Result Validity and Denominators}
\label{sec:result-validity}

To explain why Pass@2 values need not be exact multiples of $1/144$, we audit valid denominators and exclusion causes. Table~\ref{tab:main_results} nominally covers 144 environment pairs per cell. Across its method and model entries, the valid denominators average 139.20 (standard deviation 4.67) because runs with no complete, executable solution are excluded. Four model-intrinsic failure types are tracked: \textbf{output truncation}, when the response reaches the token limit before a complete solution; \textbf{parse failure}, when no solution can be extracted from the required code-block format; \textbf{sandbox runtime crash}, when candidate execution terminates without a valid score; and \textbf{context overflow}, when accumulated prompt content exceeds the model context window. For example, Qwen3-4B + Reflexion has denominator 140: two output truncations, one parse failure, zero sandbox runtime crashes, and one context overflow. These exclusions are recorded before aggregation and do not change the reported method ordering.

\subsection{Run Reliability and Cost Tradeoff}
\label{sec:run-reliability}

Table~\ref{tab:run_reliability} evaluates whether two stochastic runs support reproducible conclusions. On the full benchmark, 86.6\% pass/fail agreement, a zero median score difference, and stronger rank agreement at larger scales suggest that most outcomes are not run-specific. Adding a third Kinematics run changes only 5.1\% of pair-level pass decisions and leaves mean scores closely aligned, indicating limited marginal value from another repeat. The lower 8B rank correlation reflects a near tie among ACE, Reflexion, and ToT, which are separated by at most 1.1 score points, rather than broad instability. Since a full additional run costs approximately 2,100 A100 GPU-hours, two runs provide a practical balance between reliability and cost.

\begin{table}[t]
  \centering
  \small
  \begin{tabular}{lcc}
    \toprule
    \textbf{Model} & \textbf{Pass@2 (\%)} & \textbf{Score@2} \\
    \midrule
    Qwen3-4B  & 11.3 & 17.2 \\
    Qwen3-8B  & 12.8 & 16.8 \\
    Qwen3-14B & 18.3 & 21.7 \\
    \bottomrule
  \end{tabular}
  \caption{Vanilla from-scratch results by model, computed over eligible task--environment instances.}
  \label{tab:from-scratch-vanilla}
  \vspace{-0.2cm}
\end{table}

\begin{table}[t]
  \centering
  \small
  \begin{tabular}{lcc}
    \toprule
    \textbf{Target environment} & \textbf{Pass@2 (\%)} & \textbf{Score@2} \\
    \midrule
    Source     & 32.3 & 30.8 \\
    Mutated-1 & 13.4 & 18.5 \\
    Mutated-2 & 10.9 & 16.6 \\
    Mutated-3 & 11.2 & 15.3 \\
    Mutated-4 & 6.4 & 11.6 \\
    \bottomrule
  \end{tabular}
  \caption{Vanilla from-scratch results by target environment, computed over eligible task--environment instances.}
  \label{tab:from-scratch-environment}
  \vspace{-0.2cm}
\end{table}

\subsection{Larger-Model Results}
\label{sec:model-comparison-details}

To broaden frontier-model coverage at manageable cost, we compare six frontier LLMs with Qwen3-4B/8B/14B on the 24-pair Statics subset using Vanilla and a five-attempt budget. Because this budget differs from the main protocol, we use the study only for within-setting model comparison.

Figure~\ref{fig:bar_large_baseline_model} reveals a sharp capability gap under the five-attempt budget. Qwen3-4B/8B/14B and MiniMax-M2.7 achieve 0\% Pass@2, although Qwen3's Score@2 rises from 0.0 to 6.4 with scale. Larger Qwen3 models therefore make more partial progress but still fail to complete a design within the short interaction horizon. Among models with nonzero success, Score@2 and Pass@2 yield the same ranking: GPT-5.5 $>$ Gemini-3.1-Pro $>$ Claude-Opus-4.7 $>$ DeepSeek-V4-Pro $>$ Kimi-K2.6. GPT-5.5 leads with 36.4\% Pass@2 and 70.5 Score@2, yet still fails 63.6\% of pairs. Stronger base models thus improve both progress and success probability, but do not make short-budget adaptation reliable.

\section{Supplementary Analysis}
\label{sec:supplementary-analysis}

\subsection{Convergence Analysis}

To test how interaction budget affects model scaling, we evaluate Pass@2@$k$ and Score@2@$k$ after truncating each trajectory to $k$ attempts. Figure~\ref{fig:discovery_by_model} reports these metrics aggregated across all methods. No model succeeds with only one attempt ($k=1$). Performance then rises monotonically with budget, and the gap between model scales widens with more attempts, suggesting that larger models may extract more value from additional interaction.

\subsection{Per-Attempt Similarity}
\label{sec:per-attempt-similarity}

To probe score-retrieval anchoring over time, we compare adjacent-attempt Jaccard similarity for Vanilla and ACE in one Qwen3-14B Stage~4 case study. Table~\ref{tab:per_attempt_k06} shows that mean similarity rises from the first five to the last five transitions for both Vanilla (0.52$\to$0.75) and ACE (0.69$\to$0.92), with ACE remaining higher overall (global mean 0.77 vs.\ 0.68). ACE submits identical consecutive code at transitions 8 and 20. Vanilla instead alternates between near-fixation (0.963 at transition 12) and renewed exploration (0.500 at transition 18). This trace suggests progressive convergence in both methods and stronger score-retrieval anchoring in ACE.

\subsection{Category-Level Analysis}
\label{sec:category-level-analysis}

\begin{figure}[t]
\centering
\includegraphics[width=\columnwidth]{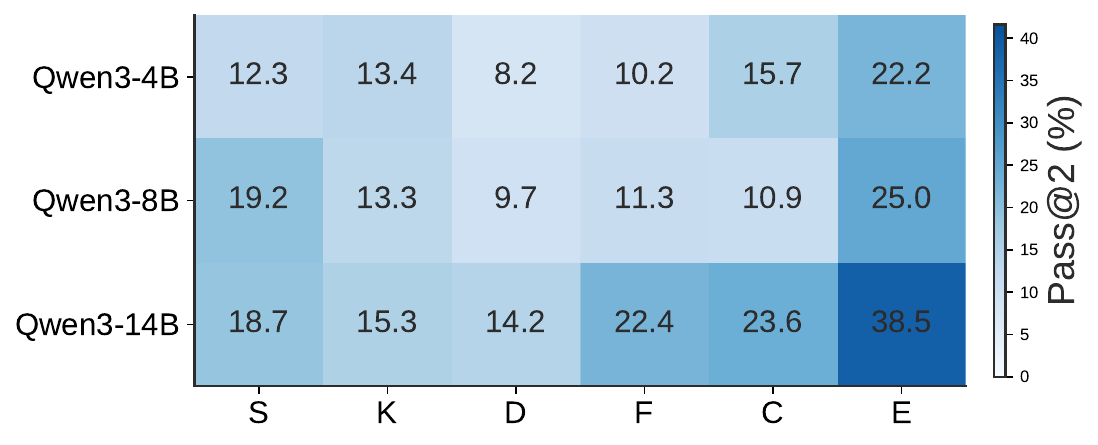}
\caption{Model $\times$ category Pass@2 heatmap (all methods aggregated).}
\label{fig:heatmap_model_category_pass}
\end{figure}

Figure~\ref{fig:heatmap_model_category_pass} reports Pass@2 aggregated across methods by model and category. A stable difficulty hierarchy emerges: \textbf{Exotic Physics} is consistently the most solvable category, while \textbf{Dynamics} remains the hardest across all model scales. Model scaling gains vary sharply by category: the 4B$\to$14B improvement is largest in \textbf{Exotic Physics} (+16.3) and \textbf{Fluid} (+12.2), and smallest in \textbf{Kinematics} (+1.9).

Error patterns are strongly category-specific. \textbf{Category~1 (Statics, S)} tests equilibrium reasoning. Agents explore diversely (radicality 0.311) but show 34.1\% Exploration without convergence. Because ``almost balanced'' structures still collapse, iterative refinement lacks a gradient. \textbf{Category~2 (Kinematics, K)} tests linkage topology design. The smallest scaling gain (+1.9) and 44.5\% Budget Exhaustion suggest that combinatorial search, rather than reasoning depth, may be the main bottleneck. \textbf{Category~3 (Dynamics, D)} tests coupled energy, momentum, and timing. All three trajectory failures co-occur (30.6\% Fixation, 30.1\% Stagnation, 28.2\% Exploration) because improving one constraint often breaks another. \textbf{Category~4 (Granular/Fluid, F)} tests many-body interaction reasoning. The unique 20.3\% Constraint Violation rate may reflect failure to anticipate emergent particle behavior from code alone. \textbf{Category~5 (Control, C)} tests closed-loop dynamics. Extreme Fixation (58.2\%, radicality 0.153) may reflect the difficulty of isolating which control parameter failed from aggregate feedback. \textbf{Category~6 (Exotic Physics, E)} tests survival under unfamiliar regimes. Agents either fail early (19.3\% Structural Failure) or converge quickly (Budget Exhaustion 1.1\%). Because survival is coarser than precision, the budget is rarely the bottleneck.

\section{Prompts}
\label{sec:prompts}

This section documents the key prompts used during benchmark construction, corresponding to the auditing, feedback design, and difficulty escalation pipelines described in~\S\ref{sec:construction} and Appendix~\S\ref{sec:construction-details}.

\subsection{Module Auditing Prompt}

The following prompt is issued to an LLM for automated auditing of each task directory (see~\S\ref{sec:module-audit}):

\begin{tcolorbox}[
    enhanced, breakable, colback=blue!3!white, colframe=blue!15!white,
    boxrule=0.5pt, arc=2mm, drop fuzzy shadow,
    left=12pt, right=12pt, top=8pt, bottom=8pt,
]
\small\textbf{Module Auditing Prompt.}
\footnotesize\ttfamily\raggedright

\#\# Objective \\
Conduct a strict, exhaustive audit of the current task directory. \\
All file paths must be relative to \texttt{.}. \\
~\\
\#\# Anti-Laziness Rule \\
Do NOT stop after finding 1 or 2 errors. Provide an EXHAUSTIVE, \\
line-by-line enumeration of EVERY SINGLE violation. \\
~\\
\#\# Variable Classification \\
~\\
\textbf{1. CONSTRAINT: MUST have numeric value in prompt.py} \\
Variables defining absolute maxima, minima, or failure thresholds \\
required to solve the task. Examples: max structure mass, \\
max joint torque, gate positions, target coordinates. \\
Invisible constraints STILL need numeric values in prompt.py. \\
~\\
\textbf{2. INVISIBLE NON-CONSTRAINT: NEVER numeric in prompt.py} \\
Background physics that cannot be visually observed. If ANY \\
numeric value appears in prompt.py, DELETE it. Affected: \\
gravity, linear\textunderscore damping, angular\textunderscore damping, wind\textunderscore amplitude, \\
drain\textunderscore velocity\textunderscore factor, slip\textunderscore backward\textunderscore force. \\
Qualitative descriptions allowed, but numeric values prohibited. \\
~\\
\textbf{3. VISIBLE VARIABLE: MUST have numeric value in prompt.py} \\
Observable physical properties: gate positions, target zone, \\
initial craft position. \\
~\\
\#\# Audit Steps \\
Step 1: Cross-Module Consistency: check physics across all \\
modules (env, eval, feedback, prompt, stages, renderer). \\
Step 2: Invisible Non-Constraint Sweep: delete any numeric \\
value of invisible variables from prompt.py. \\
Step 3: Constraint Completeness: every hardcoded constraint \\
in environment.py must appear in prompt.py. \\
Step 4: Mutation Sync: check "(originally OLD\textunderscore VALUE)" \\
format for constraint/visible changes in stages.py. \\
Step 5: UNIFORM\textunderscore SUFFIX Tone: suffix must list only generic \\
warnings, never specific values or directions. \\
Step 6: Runtime Pipeline Check: run full cross-mutation \\
pipeline with mock model across all 4 stages.
\end{tcolorbox}

\subsection{Feedback Optimization Prompts}

The feedback pipeline (see~\S\ref{sec:feedback-opt}) uses a two-phase LLM workflow.

\noindent\textbf{Phase 1: Forensic Analysis.}

\begin{tcolorbox}[
    enhanced, breakable, colback=blue!3!white, colframe=blue!15!white,
    boxrule=0.5pt, arc=2mm, drop fuzzy shadow,
    left=12pt, right=12pt, top=8pt, bottom=8pt,
]
\small\textbf{Phase 1: Forensic Analysis.}
\footnotesize\ttfamily\raggedright

You are a physics simulation forensic analyst. Analyze the \\
execution logs to determine why an LLM agent failed. \\
~\\
Six Diagnostic Dimensions: \\
\begin{itemize}[leftmargin=*, nosep, labelsep=5pt]
  \item Temporal Chronology: ordered failure timeline with step \\
    numbers and positions.
  \item Spatial Margins: every measurement paired with limit and margin.
  \item Load Distribution: components ranked by stress percentage.
  \item Energy Flow: stored $\rightarrow$ delivered $\rightarrow$ losses (dynamics tasks).
  \item Constraint Profile: ALL constraints with PASS/FAIL + margins.
  \item Numerical Health: flag NaN, Inf, extreme velocities.
\end{itemize}
List the TOP 3--5 missing diagnostics that would have unlocked \\
success, and specify how each can be computed from the sandbox.
\end{tcolorbox}

\noindent\textbf{Phase 2: Feedback Implementation.}

\begin{tcolorbox}[
    enhanced, breakable, colback=blue!3!white, colframe=blue!15!white,
    boxrule=0.5pt, arc=2mm, drop fuzzy shadow,
    left=12pt, right=12pt, top=8pt, bottom=8pt,
]
\small\textbf{Phase 2: Feedback Implementation.}
\footnotesize\ttfamily\raggedright

Implement this task's \texttt{feedback.py} to produce forensic- \\
quality diagnostic reports covering all six dimensions. \\
~\\
\textbf{Rules:} Every limit from metrics.get(). Never hardcode. \\
thresholds. Report margins (distance to limit), not raw values. \\
Order events chronologically. Sort stress data by severity. \\
NEVER prescribe engineering fixes: identify WHAT failed and \\
BY WHAT MARGIN, never "you should...". NEVER modify existing \\
defaults or pass/fail logic outside feedback.py. \\
~\\
\textbf{Task-Type Guidance:} \\
S\textunderscore{} (Statics): joint failure cascade, spatial margins, load. \\
K\textunderscore{} (Kinematics): distance vs target, joint angle extents. \\
D\textunderscore{} (Dynamics): energy chain, loss breakdown, trajectory margins. \\
F\textunderscore{} (Granular/Fluid): leakage vs limit, containment timeline. \\
C\textunderscore{} (Control): control error per zone, stability margins. \\
E\textunderscore{} (Exotic): ALL six dimensions, unusual physics.
\end{tcolorbox}

\subsection{Difficulty Escalation Prompt}

The following prompt is issued for each environment pair requiring escalation (see~\S\ref{sec:diff-escalation}):

\begin{tcolorbox}[
    enhanced, breakable, colback=blue!3!white, colframe=blue!15!white,
    boxrule=0.5pt, arc=2mm, drop fuzzy shadow,
    left=12pt, right=12pt, top=8pt, bottom=8pt,
]
\small\textbf{Difficulty Escalation Prompt.}
\footnotesize\ttfamily\raggedright

Escalate the difficulty of this target environment stage. \\
Goal: make it as hard as possible while remaining solvable by \\
the stage-specific reference solution. \\
~\\
\textbf{Critical Constraints:} \\
$\bullet$ Only mutate existing variables, with no new physics. \\
$\bullet$ Only modify \texttt{stages.py} and \texttt{agent.py}. \\
$\bullet$ Source reference MUST pass the source environment but FAIL on the target environment. \\
$\bullet$ UNIFORM\textunderscore SUFFIX identical across ALL 4 stages, generically. \\
~\\
\textbf{Verification:} \\
1. Read all task modules. \\
2. Escalate target stage in \texttt{stages.py} and update \texttt{agent.py}. \\
3. Update UNIFORM\textunderscore SUFFIX across all stages if needed. \\
4. Run \texttt{test\textunderscore reference\textunderscore solutions.py}: source passes $\mathcal{E}_0$ \\
\hspace{1em}and fails ALL 4 target environments. \\
5. Run mock evaluation pipeline to catch runtime errors.
\end{tcolorbox}


\section{Full Task Specifications}
\label{sec:task-details}

Tables~\ref{tab:cat1}--\ref{tab:cat6} provide detailed specifications for all 36 tasks, including the source task description and the physical mutations applied in each of the four target environments.

\onecolumn

\begin{longtable}{lp{6.8cm}p{7.5cm}}
  \toprule
  \textbf{ID} & \textbf{Stage} & \textbf{Mutations} \\
  \midrule
  \endfirsthead
  \toprule
  \textbf{ID} & \textbf{Stage} & \textbf{Mutations} \\
  \midrule
  \endhead
  \multicolumn{3}{p{\textwidth}}{\textbf{S\_01: Bridge for vehicle crossing}} \\
  & Stage-1: Brittle Material & \begin{minipage}[t]{7.3cm}\raggedright \textbullet\ max\_structure\_mass = 492.0 \\ \textbullet\ joint\_max\_force = 40.0 \\ \textbullet\ anchor\_max\_force = 50.0 \\ \textbullet\ joint\_max\_torque = 0.1 \\ \textbullet\ anchor\_max\_torque = 0.1\end{minipage} \\
  \cmidrule{2-3}
  & Stage-2: Paper Joints & \begin{minipage}[t]{7.3cm}\raggedright \textbullet\ max\_structure\_mass = 1000.0 \\ \textbullet\ joint\_max\_force = 5.5 \\ \textbullet\ anchor\_max\_force = 15.0\end{minipage} \\
  \cmidrule{2-3}
  & Stage-3: The Vortex Gorge & \begin{minipage}[t]{7.3cm}\raggedright \textbullet\ gap\_width = 20.0 \\ \textbullet\ max\_structure\_mass = 550.0 \\ \textbullet\ gravity = (0, -20.0) \\ \textbullet\ wind\_force = (-20.0, -4.0) \\ \textbullet\ joint\_max\_torque = 0.2 \\ \textbullet\ anchor\_max\_torque = 0.2\end{minipage} \\
  \cmidrule{2-3}
  & Stage-4: Abyssal Crossing & \begin{minipage}[t]{7.3cm}\raggedright \textbullet\ gap\_width = 26.0 \\ \textbullet\ max\_structure\_mass = 330.0 \\ \textbullet\ gravity = (0, -28.0) \\ \textbullet\ wind\_force = (-45.0, -8.0) \\ \textbullet\ joint\_max\_force = 40.0 \\ \textbullet\ anchor\_max\_force = 60.0 \\ \textbullet\ joint\_max\_torque = 80.0 \\ \textbullet\ anchor\_max\_torque = 120.0\end{minipage} \\
  \midrule
  \multicolumn{3}{p{\textwidth}}{\textbf{S\_02: Tower resisting seismic vibration}} \\
  & Stage-1: The Brittle Foundation & \begin{minipage}[t]{7.3cm}\raggedright \textbullet\ max\_joint\_torque = 5000.0 \\ \textbullet\ max\_joint\_force = 7000.0\end{minipage} \\
  \cmidrule{2-3}
  & Stage-2: The Glass Skeleton & \begin{minipage}[t]{7.3cm}\raggedright \textbullet\ max\_joint\_torque = 5000.0\end{minipage} \\
\end{longtable}

\begin{longtable}{lp{6.8cm}p{7.5cm}}
  \toprule
  \textbf{ID} & \textbf{Stage} & \textbf{Mutations} \\
  \midrule
  \endfirsthead
  \toprule
  \textbf{ID} & \textbf{Stage} & \textbf{Mutations} \\
  \midrule
  \endhead
  \multicolumn{3}{p{\textwidth}}{\textbf{K\_01: Motor-driven 2D walker}} \\
  & Stage-1: Near-Zero Structure Mass Budget & \begin{minipage}[t]{7.3cm}\raggedright \textbullet\ max\_structure\_mass = 0.8\end{minipage} \\
  \cmidrule{2-3}
  & Stage-2: Near-Zero Ground Traction + Restricted Joints & \begin{minipage}[t]{7.3cm}\raggedright \textbullet\ ground\_friction = 0.02 \\ \textbullet\ default\_joint\_lower\_limit = -0.2617993877991494 \\ \textbullet\ default\_joint\_upper\_limit = 0.2617993877991494\end{minipage} \\
  \cmidrule{2-3}
  & Stage-3: Dissipative Low-Traction Walking Crown & \begin{minipage}[t]{7.3cm}\raggedright \textbullet\ ground\_friction = 0.015 \\ \textbullet\ max\_structure\_mass = 1.5 \\ \textbullet\ gravity = (0, -17.0) \\ \textbullet\ max\_body\_friction = 0.2 \\ \textbullet\ default\_joint\_lower\_limit = -0.2243994752564138 \\ \textbullet\ default\_joint\_upper\_limit = 0.2243994752564138 \\ \textbullet\ linear\_damping = 4.0 \\ \textbullet\ angular\_damping = 4.0\end{minipage} \\
  \cmidrule{2-3}
  & Stage-4: Maximum Difficulty — All Variables Beyond Stage-3 Extremes & \begin{minipage}[t]{7.3cm}\raggedright \textbullet\ ground\_friction = 0.01 \\ \textbullet\ max\_structure\_mass = 1.2 \\ \textbullet\ gravity = (0, -20.0) \\ \textbullet\ max\_body\_friction = 0.008 \\ \textbullet\ default\_joint\_lower\_limit = -0.20943951023931953 \\ \textbullet\ default\_joint\_upper\_limit = 0.20943951023931953 \\ \textbullet\ linear\_damping = 18.0 \\ \textbullet\ angular\_damping = 18.0\end{minipage} \\
  \midrule
  \multicolumn{3}{p{\textwidth}}{\textbf{K\_02: Vertical wall climber}} \\
  & Stage-1: Fragile Structural Integrity & \begin{minipage}[t]{7.3cm}\raggedright \textbullet\ build\_zone\_y\_max = 5.0 \\ \textbullet\ max\_joint\_force = 1.5 \\ \textbullet\ max\_joint\_torque = 3.0\end{minipage} \\
\end{longtable}

\begin{longtable}{lp{6.8cm}p{7.5cm}}
  \toprule
  \textbf{ID} & \textbf{Stage} & \textbf{Mutations} \\
  \midrule
  \endfirsthead
  \toprule
  \textbf{ID} & \textbf{Stage} & \textbf{Mutations} \\
  \midrule
  \endhead
  \multicolumn{3}{p{\textwidth}}{\textbf{D\_01: Projectile launcher}} \\
  & Stage-1: The Dense Atmosphere & \begin{minipage}[t]{7.3cm}\raggedright \textbullet\ linear\_damping = 2.5 \\ \textbullet\ angular\_damping = 2.5\end{minipage} \\
  \cmidrule{2-3}
  & Stage-2: The Distant Target & \begin{minipage}[t]{7.3cm}\raggedright \textbullet\ target\_x\_min = 50.0 \\ \textbullet\ target\_x\_max = 55.0 \\ \textbullet\ target\_y\_min = 2.0 \\ \textbullet\ target\_y\_max = 5.0\end{minipage} \\
  \cmidrule{2-3}
  & Stage-3: Heavy World and Drag & \begin{minipage}[t]{7.3cm}\raggedright \textbullet\ gravity = (0, -15.0) \\ \textbullet\ linear\_damping = 1.5 \\ \textbullet\ angular\_damping = 1.5\end{minipage} \\
  \cmidrule{2-3}
  & Stage-4: Extreme Range and Conditions & \begin{minipage}[t]{7.3cm}\raggedright \textbullet\ target\_x\_min = 52.0 \\ \textbullet\ target\_x\_max = 57.0 \\ \textbullet\ target\_y\_min = 2.0 \\ \textbullet\ target\_y\_max = 5.0 \\ \textbullet\ gravity = (0, -18.0) \\ \textbullet\ linear\_damping = 1.2 \\ \textbullet\ angular\_damping = 1.2\end{minipage} \\
  \midrule
  \multicolumn{3}{p{\textwidth}}{\textbf{D\_02: Jumping mechanism}} \\
  & Stage-1: Viscous Void & \begin{minipage}[t]{7.3cm}\raggedright \textbullet\ linear\_damping = 2.0\end{minipage} \\
  \cmidrule{2-3}
  & Stage-2: Alternating Passages & \begin{minipage}[t]{7.3cm}\raggedright \textbullet\ slot1\_floor = 4.8 \\ \textbullet\ slot1\_ceil = 5.7 \\ \textbullet\ slot3\_floor = 15.0 \\ \textbullet\ slot3\_ceil = 15.9 \\ \textbullet\ slot2\_floor = 4.4 \\ \textbullet\ slot2\_ceil = 5.3\end{minipage} \\
\end{longtable}

\begin{longtable}{lp{6.8cm}p{7.5cm}}
  \toprule
  \textbf{ID} & \textbf{Stage} & \textbf{Mutations} \\
  \midrule
  \endfirsthead
  \toprule
  \textbf{ID} & \textbf{Stage} & \textbf{Mutations} \\
  \midrule
  \endhead
  \multicolumn{3}{p{\textwidth}}{\textbf{F\_01: Dam blocking water flow}} \\
  & Stage-1: Low weld ceiling (threshold physics) & \begin{minipage}[t]{7.3cm}\raggedright \textbullet\ joint\_break\_force = 41000.0\end{minipage} \\
  \cmidrule{2-3}
  & Stage-2: Elastic reservoir granules & \begin{minipage}[t]{7.3cm}\raggedright \textbullet\ fluid\_particle\_restitution = 0.78\end{minipage} \\
  \cmidrule{2-3}
  & Stage-3: Bounce apocalypse + wide squeeze + fast debris + heavy world & \begin{minipage}[t]{7.3cm}\raggedright \textbullet\ joint\_break\_force = 50000.0 \\ \textbullet\ joint\_break\_consecutive\_steps = 2 \\ \textbullet\ downstream\_wall\_amplitude = 1.0 \\ \textbullet\ downstream\_wall\_phase\_divisor = 30.0 \\ \textbullet\ fluid\_particle\_restitution = 0.95 \\ \textbullet\ debris\_linear\_velocity\_x = 3.2 \\ \textbullet\ max\_leakage\_rate = 0.004 \\ \textbullet\ gravity = (0, -10.5)\end{minipage} \\
  \cmidrule{2-3}
  & Stage-4: Apocalypse: every physics dial at breaking point & \begin{minipage}[t]{7.3cm}\raggedright \textbullet\ joint\_break\_force = 50000.0 \\ \textbullet\ joint\_break\_consecutive\_steps = 2 \\ \textbullet\ downstream\_wall\_amplitude = 1.0 \\ \textbullet\ downstream\_wall\_phase\_divisor = 20.0 \\ \textbullet\ fluid\_particle\_restitution = 0.97 \\ \textbullet\ fluid\_particle\_friction = 0.02 \\ \textbullet\ debris\_linear\_velocity\_x = 4.0 \\ \textbullet\ earthquake\_impulse\_x = 0.5 \\ \textbullet\ upward\_surge\_impulse\_y = 1.5 \\ \textbullet\ max\_structure\_mass = 130.0 \\ \textbullet\ backward\_slosh\_impulse\_x = -0.9 \\ \textbullet\ max\_leakage\_rate = 0.005 \\ \textbullet\ surge\_impulses = [1.0, 1.2, 1.3, 1.5, 1.6, 1.8, 1.9, 2.1, 2.2] \\ \textbullet\ gravity = (0, -11.5)\end{minipage} \\
  \midrule
  \multicolumn{3}{p{\textwidth}}{\textbf{F\_02: Amphibious vehicle}} \\
  & Stage-1: The Sluggish Dawn & \begin{minipage}[t]{7.3cm}\raggedright \textbullet\ thrust\_cooldown\_steps = 200\end{minipage} \\
\end{longtable}

\begin{longtable}{lp{6.8cm}p{7.5cm}}
  \toprule
  \textbf{ID} & \textbf{Stage} & \textbf{Mutations} \\
  \midrule
  \endfirsthead
  \toprule
  \textbf{ID} & \textbf{Stage} & \textbf{Mutations} \\
  \midrule
  \endhead
  \multicolumn{3}{p{\textwidth}}{\textbf{C\_01: Inverted pendulum cart-pole}} \\
  & Stage-1: Curriculum stage 1 & \begin{minipage}[t]{7.3cm}\raggedright \textbullet\ track\_center\_x = 50.0 \\ \textbullet\ pole\_start\_angle = 0.1658062789394612 \\ \textbullet\ sensor\_delay\_angle\_steps = 10 \\ \textbullet\ sensor\_delay\_omega\_steps = 10\end{minipage} \\
  \cmidrule{2-3}
  & Stage-2: Curriculum stage 2 & \begin{minipage}[t]{7.3cm}\raggedright \textbullet\ track\_center\_x = 50.0 \\ \textbullet\ cart\_force\_limit\_newtons = 0.5 \\ \textbullet\ max\_steps = 2000 \\ \textbullet\ gravity = 12.0 \\ \textbullet\ sensor\_delay\_angle\_steps = 10 \\ \textbullet\ sensor\_delay\_omega\_steps = 10 \\ \textbullet\ pole\_start\_angle = 0.003 \\ \textbullet\ safe\_half\_range = 0.1\end{minipage} \\
  \cmidrule{2-3}
  & Stage-3: Curriculum stage 3 & \begin{minipage}[t]{7.3cm}\raggedright \textbullet\ track\_center\_x = 50.0 \\ \textbullet\ gravity = 22.0 \\ \textbullet\ sensor\_delay\_angle\_steps = 3 \\ \textbullet\ sensor\_delay\_omega\_steps = 3 \\ \textbullet\ cart\_force\_limit\_newtons = 4.0 \\ \textbullet\ pole\_mass = 5.0 \\ \textbullet\ cart\_mass = 3.0 \\ \textbullet\ safe\_half\_range = 0.2 \\ \textbullet\ pole\_length = 0.7 \\ \textbullet\ max\_steps = 1000 \\ \textbullet\ pole\_start\_angle = 0.0004\end{minipage} \\
  \cmidrule{2-3}
  & Stage-4: Curriculum stage 4 & \begin{minipage}[t]{7.3cm}\raggedright \textbullet\ track\_center\_x = 50.0 \\ \textbullet\ gravity = 35.0 \\ \textbullet\ sensor\_delay\_angle\_steps = 3 \\ \textbullet\ sensor\_delay\_omega\_steps = 3 \\ \textbullet\ cart\_force\_limit\_newtons = 50.0 \\ \textbullet\ pole\_mass = 12.0 \\ \textbullet\ cart\_mass = 2.0 \\ \textbullet\ safe\_half\_range = 0.15 \\ \textbullet\ pole\_length = 0.5 \\ \textbullet\ max\_steps = 500 \\ \textbullet\ pole\_start\_angle = 5e-05 \\ \textbullet\ cart\_rail\_center\_y = 6.0\end{minipage} \\
\end{longtable}

\begin{longtable}{lp{6.8cm}p{7.5cm}}
  \toprule
  \textbf{ID} & \textbf{Stage} & \textbf{Mutations} \\
  \midrule
  \endfirsthead
  \toprule
  \textbf{ID} & \textbf{Stage} & \textbf{Mutations} \\
  \midrule
  \endhead
  \multicolumn{3}{p{\textwidth}}{\textbf{E\_01: Surviving reversing gravity}} \\
  & Stage-1: Altered Field I & \begin{minipage}[t]{7.3cm}\raggedright \textbullet\ arena\_y\_max = 20.0 \\ \textbullet\ build\_zone\_y\_max = 18.0 \\ \textbullet\ gravity = gravity\_resonant\_violent(t) \\ \textbullet\ joint\_force\_limit = 1.0\end{minipage} \\
  \cmidrule{2-3}
  & Stage-2: Altered Field II & \begin{minipage}[t]{7.3cm}\raggedright \textbullet\ arena\_y\_max = 9.0 \\ \textbullet\ build\_zone\_y\_max = 8.5 \\ \textbullet\ friction = 0.0 \\ \textbullet\ gravity = gravity\_cataclysm(t) \\ \textbullet\ beam\_density\_scale = 200.0 \\ \textbullet\ linear\_damping = -6.0 \\ \textbullet\ angular\_damping = -4.0 \\ \textbullet\ joint\_force\_limit = 0.001 \\ \textbullet\ max\_beam\_count = 1 \\ \textbullet\ max\_structure\_mass = 0.02\end{minipage} \\
  \cmidrule{2-3}
  & Stage-3: Altered Field III & \begin{minipage}[t]{7.3cm}\raggedright \textbullet\ arena\_y\_max = 9.5 \\ \textbullet\ build\_zone\_y\_max = 9.0 \\ \textbullet\ friction = 0.0 \\ \textbullet\ gravity = gravity\_vortex\_singularity(t) \\ \textbullet\ beam\_density\_scale = 150.0 \\ \textbullet\ linear\_damping = -6.0 \\ \textbullet\ angular\_damping = -4.0 \\ \textbullet\ joint\_force\_limit = 0.15 \\ \textbullet\ max\_beam\_count = 2 \\ \textbullet\ max\_structure\_mass = 0.3\end{minipage} \\
  \cmidrule{2-3}
  & Stage-4: Altered Field IV & \begin{minipage}[t]{7.3cm}\raggedright \textbullet\ arena\_y\_max = 9.0 \\ \textbullet\ build\_zone\_y\_max = 8.5 \\ \textbullet\ friction = 0.0 \\ \textbullet\ gravity = gravity\_cataclysm(t) \\ \textbullet\ beam\_density\_scale = 300.0 \\ \textbullet\ linear\_damping = -6.0 \\ \textbullet\ angular\_damping = -4.0 \\ \textbullet\ joint\_force\_limit = 0.01 \\ \textbullet\ max\_beam\_count = 1 \\ \textbullet\ max\_structure\_mass = 0.005\end{minipage} \\
  \midrule
  \multicolumn{3}{p{\textwidth}}{\textbf{E\_02: Thruster flyer in fluid}} \\
\end{longtable}

\twocolumn
\end{document}